\documentclass[]{mosi}

\usepackage{microtype}
\usepackage{colortbl}
\usepackage{tabularx}
\usepackage{fvextra}
\usepackage{needspace}

\graphicspath{{../}{./}}

\definecolor{agentrow}{RGB}{245,248,252}
\definecolor{solverrow}{RGB}{250,246,238}
\definecolor{headerrow}{RGB}{235,235,235}
\definecolor{zerocallcell}{RGB}{255,229,229}
\definecolor{metriclabelcell}{RGB}{222,237,227}
\definecolor{agentmetriccell}{RGB}{236,246,255}
\definecolor{referencemetriccell}{RGB}{255,248,218}
\definecolor{agentbestmetriccell}{RGB}{190,222,255}
\definecolor{referencebestmetriccell}{RGB}{255,231,145}

\newcolumntype{M}{>{\columncolor{metriclabelcell}\raggedright\arraybackslash}X}
\newcolumntype{A}{>{\columncolor{agentmetriccell}}r}
\newcolumntype{R}{>{\columncolor{referencemetriccell}}r}
\newcommand{\bestagent}[1]{\cellcolor{agentbestmetriccell}\textbf{#1}}
\newcommand{\bestref}[1]{\cellcolor{referencebestmetriccell}\textbf{#1}}

\DefineVerbatimEnvironment{promptblockinner}{Verbatim}
  {fontsize=\scriptsize, frame=single, framesep=1mm, breaklines=true, breakanywhere=true, breaksymbolleft={}}
\newenvironment{promptblock}
  {\VerbatimEnvironment\par\smallskip\Needspace{10\baselineskip}\noindent\begin{promptblockinner}}
  {\end{promptblockinner}\par\smallskip}

\DefineVerbatimEnvironment{promptblockinnercompact}{Verbatim}
  {fontsize=\tiny, frame=single, framesep=0.75mm, breaklines=true, breakanywhere=true, breaksymbolleft={}}
\newenvironment{promptblockcompact}
  {\VerbatimEnvironment\par\smallskip\Needspace{6\baselineskip}\noindent\begin{promptblockinnercompact}}
  {\end{promptblockinnercompact}\par\smallskip}
\newcommand{\promptheadingcompact}[1]{\Needspace{4\baselineskip}\paragraph{#1}\mbox{}\par}

\hypersetup{
  pdftitle={AstroAgentBench: Evaluating Agentic Planning on Space Mission Planning Tasks},
  pdfauthor={Weiyi Wang, Xinchi Chen, Jingjing Gong, Xuanjing Huang, Xipeng Qiu}
}

\newcommand\plaindataformat[2][]{{\small {\checkdatafont #2}}}
\newcommand\plaindata[1]{\addtolist[]{#1}{\checkdatalist}{\plaindataformat}{\par}}
\title{AstroAgentBench: Evaluating Agentic Planning on Space Mission Planning Tasks}
\author[1,5]{Weiyi Wang}
\author[1,3,4,5,\dag]{Xinchi Chen}
\author[2,5]{Jingjing Gong}
\author[1,4]{Xuanjing Huang}
\author[1,2,3,4,5,\dag]{Xipeng Qiu}
\affiliation[1]{Institute of Trustworthy Embodied AI, Fudan University}
\affiliation[2]{Shanghai Innovation Institute}
\affiliation[3]{Shanghai Key Laboratory of Multimodal Embodied AI}
\affiliation[4]{College of Computer Science and Artificial Intelligence, Fudan University}
\affiliation[5]{OpenMOSS Team}
\plaindata{\texttt{wywang26@m.fudan.edu.cn}, \texttt{xc\_chen@fudan.edu.cn}, \texttt{jjgong@sii.edu.cn}, \texttt{xjhuang@fudan.edu.cn}, \texttt{xpqiu@fudan.edu.cn}}
\checkdata[Code]{\url{https://github.com/Mtrya/AstroAgentBench}}
\checkdata[Data]{\url{https://huggingface.co/datasets/kaupane/AstroAgentBench}}
\checkdata[Run traces]{\url{https://doi.org/10.5281/zenodo.23084446}}

\abstract{
Recent LLM-for-Space systems address mission planning, scheduling, operations support, simulator control, and autonomy, but their evaluations use different task contracts, control settings, simulators, and success criteria. We introduce AstroAgentBench, a seven-family benchmark for executable space mission planning in the domains of scheduling, observation planning, constellation design, and relay support. For each case, an agent submits a planning artifact that is checked by an external verifier for schema, timing, geometry, resources, and mission value. Results report validity and normalized scores, with comparisons to task-specific solver references. Across five LLM agent systems and 35 held-out cases, the strongest systems approach or exceed solver-reference scores on several families, while weaker systems often fail to produce high-value valid plans and even strong systems lose quality on geometric, product-level, or design-heavy tasks. Trace analyses separate two failure points: task-contract misformulation and weak solution construction. Successful runs instead calibrate agent-written implementations against verifier feedback and adapt search to case-specific structure. Ablations show that procedure injection and memory accumulation help selectively, when they supply the missing formulation, calibration, or search support.
}

\begin{document}
\maketitle
\begingroup
\def\thefootnote{\fnsymbol{footnote}}%
\footnotetext[2]{Corresponding authors.}%
\endgroup
\begin{figure*}[t]
\centering
\includegraphics[width=\textwidth]{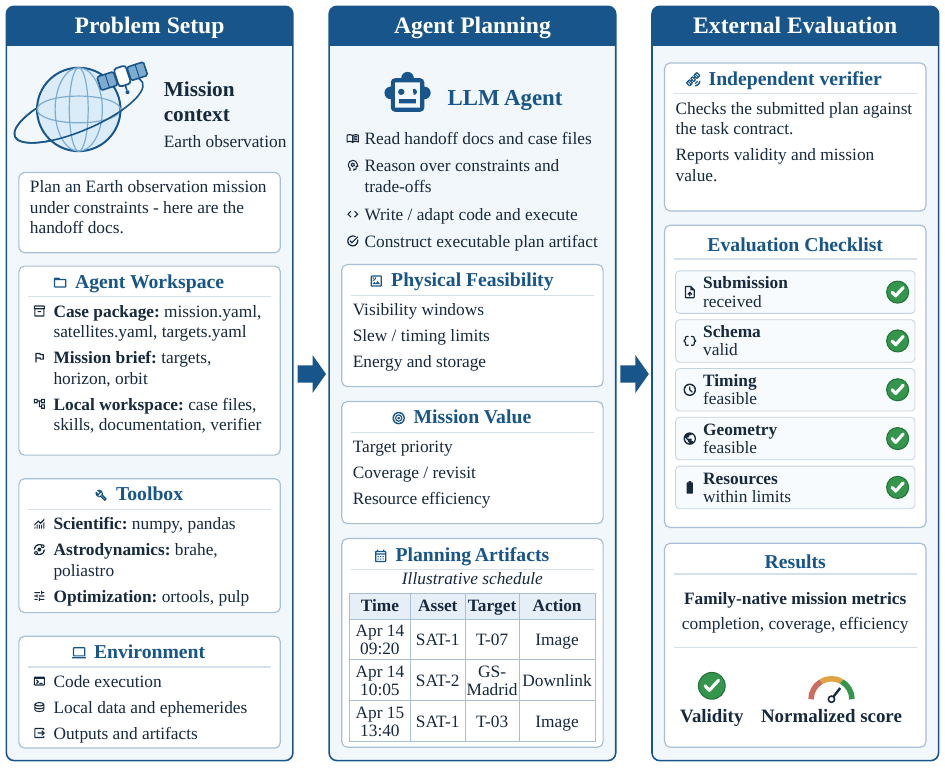}
\caption{Overview of AstroAgentBench. A case package and mission brief define the agent workspace; the agent constructs an executable planning artifact; an independent verifier checks feasibility and scores mission value.}
\label{fig:overview}
\end{figure*}

\section{Introduction}

Space mission planning is both consequential and difficult. It turns mission goals into executable decisions over spacecraft, ground assets, communication links, energy, data, and time. These decisions are coupled over long horizons and constrained by physical state, operational priorities, and limited resources. A plan that sounds plausible can still fail because it violates visibility, timing, or resource constraints. This difficulty is reflected in decades of work on AI planning, operations research, spacecraft autonomy, and satellite scheduling, which have produced specialized models and solvers for particular mission settings \citep{johnston2006automating,lemaitrea2002selecting,he2018improved,lee2024satellite}. Recent LLM-for-Space work extends the same ambition toward more flexible planning and operations support, including formation mission planning, satellite scheduling, simulated operations assistance, and orbital autonomy \citep{zhang2025large,chen2025large,campanelli_2025_llm_satellite_ops,mousist_2025_astrea}. These efforts remain fragmented across task formulations, simulators, and success criteria. General agentic planning under executable, physically grounded mission constraints remains an open evaluation problem.

Modern LLM agents have moved beyond single-turn answer generation toward systems that reason, act, and revise through interaction. Reasoning-and-acting frameworks interleave deliberation with tool use. Embodied and web agents extend this loop to persistent environments. Software agents expose repositories, command execution, and executable code as the action surface \citep{yao2022react,wangvoyager,zhouwebarena,yang2024sweagent,wang2024codeact,wang2024openhands}. Space mission planning has a similar executable surface: handoff documents, machine-readable cases, computational libraries, schedules, and checkers. This lets us ask whether an agent can translate natural-language mission goals and technical documentation into an executable planning procedure, and revise it using feedback.

We ask three questions about LLM agents in executable space mission planning. First, can they produce plans that pass external feasibility checks rather than merely describe plausible strategies? Second, when a submission is valid, how close is its quality to that of specialized solver references? Third, which system conditions explain variation in validity and scores, including verifier use, domain guidance, and accumulated experience? The distinction matters because validity alone is too weak. A plan can satisfy a schema and still observe no valuable targets, serve no demand, or cover little area. A benchmark for this setting must therefore evaluate both feasibility and graded mission value.

We introduce \textbf{AstroAgentBench}, a verifier-backed suite covering seven executable space mission-planning task families across scheduling, observation planning, constellation design, and relay support. Each case provides mission files, documentation, libraries, a solution schema, and a verifier (Figure~\ref{fig:overview}). The agent must construct a planning procedure rather than call a pre-built domain planner. By combining heterogeneous task families, external feasibility checks, graded mission value, and system-level attribution, the benchmark tests agent construction of executable plans under physical constraints.

We evaluate five LLM agent systems on AstroAgentBench and compare their outputs with per-family solver references. The main pattern is uneven capability: stronger systems often produce verifier-valid plans and sometimes approach solver-reference scores, but their success is task- and system-dependent, while weaker systems often fail before producing valid executable artifacts. The same invalid or low-value outcome can arise from different parts of the run, from task interpretation to plan construction and final submission.

We further study two forms of workspace support for solution construction. Procedure injection tests whether compact domain guidance supplies useful abstractions or brittle templates. Memory accumulation tests whether experience accumulated on training cases transfers to held-out test cases. AstroAgentBench therefore treats grounded planning as an agent-system property rather than a model-only property.

The paper makes three contributions. First, it introduces a seven-family verifier-backed benchmark for executable space mission planning, with schemas, instances, verifiers, and solver-reference comparisons. Second, it provides an empirical evaluation of contemporary LLM agent systems on this benchmark, measuring feasibility and comparing normalized scores with task-specific solver references. Third, it analyzes how trace-level failure modes, verifier use, procedure injection, and memory accumulation shape performance. AstroAgentBench gives emerging LLM-for-Space work a controlled evaluation setting grounded in physical and astrodynamic checks of feasibility and mission value.

\section{Related Work}

\subsection{LLM-Based Agents}



LLM-based agents combine language-model reasoning with tool use and environment interaction \citep{yao2022react,wangvoyager}. Coding agents extend this approach to repository navigation, code editing, and execution, with systems such as SWE-agent, CodeAct, OpenHands, and AutoCodeRover providing different interfaces and scaffolds for these activities \citep{yang2024sweagent,wang2024codeact,wang2024openhands,zhang2024autocoderover}.

\subsection{Space Mission Planning and LLM-for-Space Work}

Classical space mission planning is organized around specialized optimization problems. Deep Space Network allocation, agile satellite observation scheduling, stereo imaging, area coverage, constellation design, and satellite-network routing each define their own states, decision variables, physical constraints, and objective functions \citep{johnston2006automating,goh2022satnet,lemaitrea2002selecting,he2018improved,zezhong2023multiple,lee2024satellite,lyu2024dynamic}. This literature supplies mature solvers for individual mission-planning families, where candidate plans are meaningful only through domain-specific feasibility and value criteria.

LLM-for-Space work brings language-model systems into planning, scheduling, operations, and control workflows. Formation mission-planning work uses LLMs to decompose and organize spacecraft coordination tasks, while Earth-observation scheduling work uses multi-agent LLM systems or LLM-assisted search to design scheduling algorithms \citep{zhang2025large,chen2025large,evalns_2026_llm_satellite_scheduling}. Satellite-operations work uses LLM agents for telemetry monitoring, procedure retrieval, anomaly explanation, and human-confirmed command execution in simulated or operational settings \citep{campanelli_2025_llm_satellite_ops,esa_2024_a2i_roadmap}. Spacecraft-operation studies use LLMs or vision-language models as controllers in rendezvous and simulator environments, and on-orbit autonomy work explores LLM supervision of control policies under spacecraft resource constraints \citep{carrasco2025large,mousist_2025_astrea}. These studies span the mission lifecycle, but use different task scopes and control settings.

\subsection{Agent Benchmarks}

Agent benchmarks have expanded from symbolic planning tasks to interactive and code-grounded environments. PlanBench, planning-ability studies, and temporal-constraint benchmarks focus on action structure, state change, and goal satisfaction, while TravelPlanner studies itinerary construction under realistic travel constraints \citep{valmeekam2023planning,valmeekam2023planbench,ding2025tcp,xie2024travelplanner}. WebArena places agents in stateful digital environments where success depends on navigation, tool use, and recovery from partial observations \citep{zhouwebarena}. SWE-bench, SciCode, ScienceAgentBench, and BLADE require repository edits, executable programs, or scientific analyses that are checked by tests and programmatic metrics \citep{jimenezswe,scicode_2024,scienceagentbench_2025,gu2024blade}. These benchmarks make agent evaluation interactive and externally checked through web-state success, software tests, scientific code tests, or analysis metrics. What remains uncommon is physically grounded planning whose submitted artifact is verifiable against a domain model.

\section{AstroAgentBench}
\label{sec:benchmark}

\subsection{Overview}
\label{sec:benchmark-overview}

AstroAgentBench is a seven-family benchmark for executable space mission planning with LLM agents, spanning scheduling, imaging, and constellation design. A case gives the agent a mission instance, machine-readable files, documentation, libraries, and a required solution format. The submitted artifact is a schedule, assignment, geometric observation plan, constellation design, or contact plan that is checked after the run by an external verifier.

The suite contains 116 generated and normalized cases spanning train/test splits and seven task families. Agile Earth-observation scheduling, SatNet, and SPOT-5 center on selecting or assigning precomputed opportunities. Stereo Imaging and Regional Coverage require geometric observation plans. Revisit Constellation and Relay Constellation add constellation-design variables together with operating schedules. Table~\ref{tab:benchmark-input-scale} summarizes the per-case input scale, and Appendix~\ref{app:task-contracts} gives the per-family train/test breakdown (Table~\ref{tab:case-splits}) and the family-specific task definitions.

\begin{table}[t]
\centering
\footnotesize
\setlength{\tabcolsep}{3pt}
\begin{tabular}{>{\raggedright\arraybackslash}p{0.19\columnwidth}>{\raggedright\arraybackslash}p{0.15\columnwidth}>{\raggedright\arraybackslash}p{0.56\columnwidth}}
\hline
\textbf{Task family} & \textbf{Horizon} & \textbf{Primary input entries per case} \\
\hline
AEOSSP & 12 h & 20--28 satellites; 1,614--1,970 tasks \\
SatNet & 7 d & 257--333 requests; 2,513--3,370 view periods \\
SPOT-5 & / & 8--1,057 candidate photographs \\
Stereo & 48 h & 10--12 satellites; 121--144 targets \\
Regional & 72 h & 6--12 satellites; 10,492--18,325 grid cells \\
Revisit & 48 h & 24--32 targets \\
Relay & 96 h & 6--10 satellites; 4--8 demand windows \\
\hline
\end{tabular}
\caption{Per-case input scale in AstroAgentBench. Entries summarize the agent-facing mission objects that define each planning instance.}
\label{tab:benchmark-input-scale}
\end{table}

AstroAgentBench is related to solver-facing satellite-planning benchmarks such as SPOT-5, SatNet, AEOS-Bench, and EOS-Bench, which each focus on one planning formulation, and to broader agent benchmarks outside space mission planning. In this landscape, AstroAgentBench places seven space mission-planning families in an agent-facing setting. Appendix~\ref{app:benchmark-landscape} gives a compact comparison.

\subsection{Feasibility Constraints}
\label{sec:feasibility-constraints}

AstroAgentBench checks feasibility by reconstructing the submitted plan over the mission horizon. The verifier combines submitted decisions with the case files and simulation model, then applies the validity components in Table~\ref{tab:evaluation-layers}. Appendix~\ref{app:task-contracts} gives the family-specific validity rules, and Appendix~\ref{app:physical-models} gives the simulation and physical models.

The validity layer combines submission-format checks with physical reconstruction. File-presence and schema checks determine whether the verifier can interpret the decision object. Time, geometry, and resource checks then evaluate that object as a coupled mission plan: actions must be placed in admissible intervals, supported by the spatial configuration at those intervals, and compatible with shared assets and evolving capacities. Validity is therefore an aggregate property of the submitted plan, not only a local property of individual actions.

\begin{table*}[t]
\centering
\small
\setlength{\tabcolsep}{5pt}
\begin{tabular}{>{\raggedright\arraybackslash}p{0.18\textwidth}>{\raggedright\arraybackslash}p{0.76\textwidth}}
\hline
\textbf{Component} & \textbf{Description} \\
\hline
\multicolumn{2}{c}{\cellcolor{gray!18}\textit{Validity}} \\
Submission & The agent submits the required planning artifact before timeout. \\
Schema & The artifact parses into the family-specific decision format. \\
Time & Actions lie inside horizons, access or service windows, and required temporal gaps. \\
Geometry & The verifier recomputes visibility, pointing, range, footprint, illumination, or line of sight. \\
Resources & The plan respects concurrency limits, stateful consumables, and capacity or demand budgets. \\
\hline
\multicolumn{2}{c}{\cellcolor{gray!18}\textit{Performance}} \\
Native value & Family metrics are computed for valid artifacts, such as service, coverage, product quality, revisit gap, latency, or unmet demand. \\
Normalized score & Native values are mapped to a higher-is-better cross-family scale; missing or invalid submissions receive zero. \\
\hline
\end{tabular}
\caption{Evaluation layers in AstroAgentBench. Validity is checked before mission value is interpreted.}
\label{tab:evaluation-layers}
\end{table*}

\subsection{Benchmark Construction Pipeline}
\label{sec:construction-pipeline}

AstroAgentBench uses two construction routes. Five families are generated as new mission-planning cases, while SatNet and SPOT-5 are normalized from existing planning benchmarks. The generated route builds mission settings; the legacy route brings established instances into the same agent-facing setting.

Generated cases begin from grounded source material. Earth-observation, stereo-imaging, and regional-coverage cases sample satellites from fixed CelesTrak TLE snapshots, so orbital motion comes from real cataloged objects. Ground targets, city targets, regions, and relay endpoints are drawn from public city, site, or region libraries with global coverage. Missing sensor, agility, power, and link parameters are assigned from representative ranges.

Difficulty controls shape the candidate before physical filtering. For each generated family, they set the scale of assets and service objects, mission-horizon length or placement, geographic spread, geometry thresholds, resource budgets, and demand density. The sampled candidate contains the assets available to the agent, the service objects that create mission value, and the family-specific parameters that define successful service.

Physical filtering removes candidates that are physically empty, degenerate, unrealistic, or nearly automatic. Propagation, line-of-sight checks, access-window computation, coverage grids, lookup tables, and family-specific audits retain cases with feasible opportunities that still interact through time, geometry, and resources.

SatNet and SPOT-5 contribute published communication-scheduling and photography-selection instances. They are converted into the same evaluation shape as the generated families: the agent receives a planning instance, submits a family-specific artifact, and is scored by verifier-computed feasibility and value.

\subsection{Evaluation}
\label{sec:evaluation}

The main evaluation uses 35 held-out test cases, five from each task family. In each system-case run, the agent receives the task prompt, brief, case files, and construction materials. The submitted planning artifact is scored after the run by an external verifier.

Evaluation follows the two-layer structure in Table~\ref{tab:evaluation-layers}: validity first, then performance. The validity layer asks whether the agent completed an executable handoff under the required schema and constraints. The performance layer assesses solution quality using family-native metrics and summarizes performance with normalized scores. Agent and solver-reference results use the same family-specific scoring rules. This separates whether an agent can produce a feasible mission plan from how much case-specific value the feasible plan recovers.

For example, on one AEOSSP test case, a valid schedule from Claude Code + Claude Opus 4.6 completes 1,415 of 1,947 tasks and recovers 68.54\% of the total task weight. Combining these completion measures with turnaround time and energy consumption yields a score of 70.87; the system's five-case mean is 58.45. Appendix~\ref{app:task-contracts} gives the numerical calculation.

\subsection{Agent-System Setup}
\label{sec:agent-system-setup}

AstroAgentBench evaluates complete agent systems: a configured model, a harness, and its workspace. Systems that share a model but differ in harness, tool access, memory, or feedback channel are treated as distinct evaluated systems.

Each run is framed by an agent-facing prompt package: a task prompt, a family-specific brief, case files, and a local validation helper. During a run, agents read mission data, choose a planning abstraction, write or adapt code, execute it, inspect feedback, and revise the produced plan.

Each run uses a controlled execution environment under a fixed budget. The protocol evaluates the submitted artifact produced within that budget, without manual intervention or post-hoc repair.

\section{Experiments}
\label{sec:experiments}

We evaluate the five agent systems under a common execution budget. Each run uses the same case package, installed scientific, astrodynamics, and operations-research libraries, and a fixed two-hour wall-clock budget on 8 AMD Ryzen 7 9700X cores and 32 GB memory.

The evaluated systems are Claude Code + Claude Opus 4.6, Codex CLI + GPT-5.4, Kimi CLI + Kimi K2.6, OpenCode + MiniMax M2.7, and OpenCode + DeepSeek V4 Pro. Appendix~\ref{app:agent-workspace} gives the full environment, workspace, and system-configuration details.

We additionally report a SPOT-5 evaluation of OpenCode with self-hosted Qwen3.6-27B in Appendix~\ref{app:qwen-supplement}.

\begin{table*}[!t]
\centering
\small
\resizebox{\textwidth}{!}{%
\begin{tabular}{lcrrrrrrr}
\hline
\rowcolor{headerrow}
Method & Valid & AEOSSP & Regional & Relay & Revisit & SatNet & SPOT-5 & Stereo \\
\hline
\rowcolor{agentrow}
Claude Code + Claude Opus 4.6 & 26/35 & 58.45 & 21.22 & 12.25 & \textbf{76.50} & 60.36 & 43.96 & 19.16 \\
\rowcolor{agentrow}
Codex CLI + GPT-5.4 & \textbf{35/35} & \textbf{72.91} & 72.08 & \textbf{64.91} & 69.58 & \textbf{69.45} & \textbf{61.92} & \textbf{68.91} \\
\rowcolor{agentrow}
Kimi CLI + Kimi K2.6 & \textbf{35/35} & 72.03 & \textbf{73.15} & 57.77 & 73.36 & 66.14 & \textbf{61.92} & 0.25 \\
\rowcolor{agentrow}
OpenCode + MiniMax M2.7 & 23/35 & 36.44 & 17.73 & 0.00 & 0.00 & 37.54 & 37.47 & 0.00 \\
\rowcolor{agentrow}
OpenCode + DeepSeek V4 Pro & 34/35 & 71.44 & 35.79 & 35.67 & 70.95 & 60.23 & 60.91 & 13.00 \\
\hline
\rowcolor{solverrow}
Best solver baseline & / & \textbf{75.95} & \textbf{86.44} & 61.79 & 70.98 & 61.59 & 60.92 & \textbf{96.05} \\
\rowcolor{solverrow}
Second solver baseline & / & 71.65 & 76.26 & 58.20 & 63.33 & 55.95 & / & 92.00 \\
\hline
\end{tabular}%
}
\caption{Mean normalized score by method and task family. Scores are averaged over five held-out test cases per family; higher is better. The \textit{Valid} column counts verifier-valid agent submissions out of 35 test cases. Bold values mark the best solver-or-agent score in each family and, when different, the best agent score. For solver-baseline rows, the best and second-best solver references are selected separately for each case and then averaged within each family; SPOT-5 has only one solver-reference row.}
\label{tab:main-results}
\end{table*}

\subsection{Baselines}
\label{sec:baselines}

We compare agent systems with task-specific solver references under the same family scoring rules. Five families use implemented solvers whose outputs are checked by the benchmark verifiers; SatNet uses literature-reported results, and SPOT-5 uses archived reference solutions. For each case, we select the best and, where available, second-best reference scores, then average them within each family to obtain the solver rows in Table~\ref{tab:main-results}. Appendix~\ref{app:solver-baselines} details the methods, reference sources, and aggregation procedure.

\subsection{Main Results}
\label{sec:main-results}

Table~\ref{tab:main-results} summarizes the main result matrix. Every system has at least some verifier-valid plans, and two are valid on all 35 held-out cases. Normalized scores nevertheless remain far from saturated across several families, especially Regional Coverage and Stereo Imaging. Validity and mission value are therefore distinct measurements. With five held-out cases per family, these results provide an initial comparison of the evaluated systems. Appendix~\ref{app:score-uncertainty} reports 95\% case-bootstrap confidence intervals for the mean scores.

The score gap between agents and solver references varies across task families. On AEOSSP, the strongest agent row scores $3.04$ points below the best solver-reference row; on SPOT-5, Relay Constellation, Revisit Constellation, and SatNet, the strongest agent row scores above the corresponding reference row by $1.00$, $3.12$, $5.52$, and $7.86$ points, respectively. Regional Coverage and Stereo Imaging remain more separated: the best solver baselines average 86.44 and 96.05, while the strongest agent rows average 73.15 and 68.91.

Scores also vary widely within systems. A row that is strong on one family can leave large gaps on geometry-heavy or design-heavy families, and the family leaders differ across agents.

\section{In-Depth Analysis}
\label{sec:analysis}

\begin{figure*}[t]
\centering
\includegraphics[width=\textwidth]{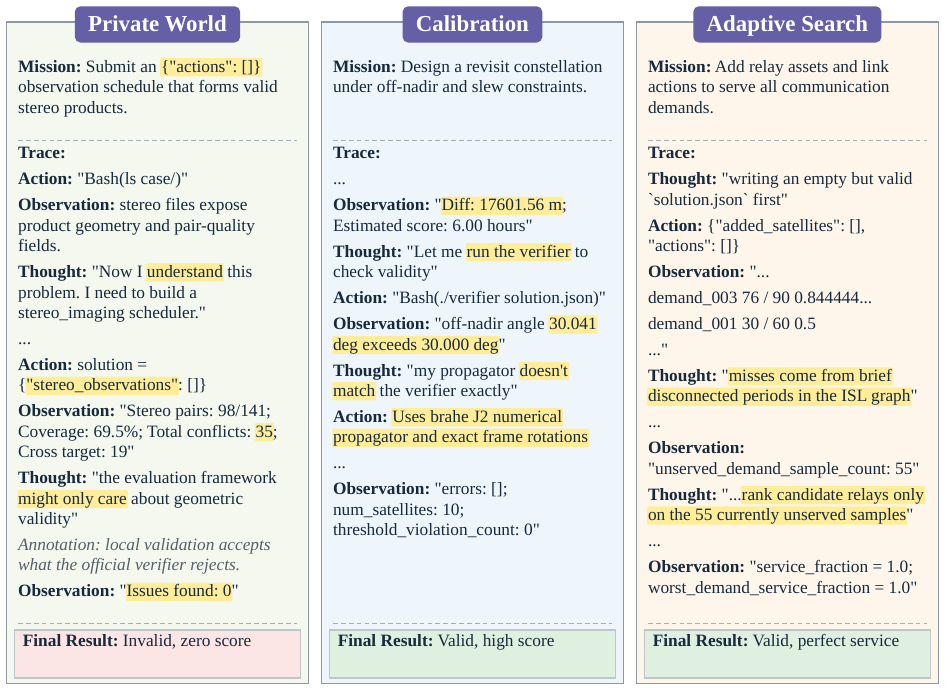}
\caption{Trace excerpts for three mechanisms in Section~\ref{sec:analysis}. Left: a Stereo Imaging run builds a self-imposed task contract, writes a product-level artifact, and accepts local validation even though the official result is invalid with zero score. Middle: a Revisit Constellation run uses verifier feedback to align propagation and frame handling with off-nadir and slew constraints. Right: a Relay Constellation run preserves a valid baseline and focuses search on unserved demand samples until service is complete.}
\label{fig:case-study-traces}
\end{figure*}

The result matrix in Section~\ref{sec:experiments} shows uneven capability rather than uniform failure. We analyze failures in task formulation and solution construction, identify two mechanisms behind strong runs, and use verifier-call patterns plus workspace-support ablations to probe those mechanisms. Figure~\ref{fig:case-study-traces} previews the trace-level evidence.

\subsection{Failure Modes}
\label{sec:what-fails}

Trace evidence separates two failure layers: formulating the task from the available materials, and constructing a high-value plan after the task has been formulated. In \textbf{task-formulation failures}, the system may read the task document but fail to integrate the physical model, coupled constraints, scored objective, and required submitted artifact. Stereo Imaging gives the most compact example. Access-window selection is only the first step: raw observations must remain valid under hard constraints, form eligible stereo products, satisfy footprint overlap, convergence, and pixel-scale checks, and preserve final validity. The observed failures break different links in this pipeline. Kimi CLI + Kimi K2.6 produces valid observations whose footprints often do not create products. Claude Code + Claude Opus 4.6 sometimes treats rich case files as a complete task definition, builds a plausible self-imposed problem, and optimizes an inferred product-level schema and hallucinated objective. Some Stereo Imaging failures also reflect ambiguity in the documented cross-track sign convention (Appendix~\ref{app:additional-case-studies}), so these outcomes measure sensitivity to the supplied task specification as well as planning capability.

In \textbf{solution-construction failures}, the system has enough of the task contract to generate candidates, but cannot reliably build, improve, and preserve a high-value executable plan. The failure appears as weak search, brittle code, late verifier use, or silent goal drift during optimization. OpenCode + MiniMax M2.7 is representative. It often contacts the task document or verifier, yet still accepts valid but zero-value relay submissions, trusts agent-written simulators after verifier disagreement, or treats feasibility as the stopping condition. OpenCode + DeepSeek V4 Pro shows a related pattern on Stereo Imaging: a run may find useful product geometry, but lose the final score by failing to preserve both product value and hard-constraint validity. Table~\ref{tab:verifier-calls} shows the aggregate counterpart for selected families: Claude Code + Claude Opus 4.6 and OpenCode + MiniMax M2.7 include several no-call cases, so those runs could not use verifier feedback to correct an agent-inferred task model or simulator during construction. Appendix~\ref{app:hallucination-metrics} quantifies the process-level counterparts of these failures across all 175 main-experiment runs: 12.6\% end with an invalid submission, 9.1\% submit a valid artifact that scores zero, and 22.3\% commit to a durable artifact before reading the root workspace contract.

\subsection{Success Mechanisms}
\label{sec:what-succeeds}

In the inspected high-scoring traces, agents often succeed by building a case-specific solver during the run and calibrating it against verifier feedback. The first mechanism is \textbf{implementation calibration}. Successful systems write their own propagation, geometry, scheduling, or routing code, but they use verifier disagreement to correct that code before trusting it. In Revisit Constellation, Claude Code + Claude Opus 4.6 first builds a close but invalid agent-written orbital model. Verifier errors expose off-nadir and slew-gap mismatches, and the run repairs its implementation around benchmark-compatible propagation, frame rotation, and transition timing. The complementary pattern appears in Table~\ref{tab:verifier-calls}: Codex CLI + GPT-5.4, Kimi CLI + Kimi K2.6, and OpenCode + DeepSeek V4 Pro call the verifier in every shown case group, giving their agent-written solvers repeated opportunities to align with that model.

\begin{table}[t]
\centering
\footnotesize
\setlength{\tabcolsep}{3.2pt}
\begin{tabular}{lrrrrr}
\hline
System & Revisit & Stereo & SatNet & Reg. & Relay \\
\hline
Claude & 13.6(0) & \cellcolor{zerocallcell}20.2(2) & \cellcolor{zerocallcell}19.0(1) & \cellcolor{zerocallcell}0.0(5) & \cellcolor{zerocallcell}1.0(4) \\
Codex & 5.8(0) & 8.8(0) & 6.0(0) & 8.2(0) & 6.4(0) \\
Kimi & 9.2(0) & 10.4(0) & 16.0(0) & 11.6(0) & 8.8(0) \\
OC+MM & \cellcolor{zerocallcell}3.0(4) & \cellcolor{zerocallcell}18.0(3) & 19.4(0) & \cellcolor{zerocallcell}25.8(1) & 19.4(0) \\
OC+DS & 6.0(0) & 47.2(0) & 10.0(0) & 18.2(0) & 12.4(0) \\
\hline
\end{tabular}
\caption{Verifier use on selected held-out task families. Each cell gives mean verifier calls per case over five cases; parentheses give no-call cases, and shading marks cells with at least one. No-call cases indicate runs without verifier feedback for calibration. Abbreviations follow Table~\ref{tab:main-results}; OC denotes OpenCode, MM/DS denote MiniMax M2.7/DeepSeek V4 Pro, and Reg. denotes Regional Coverage.}
\label{tab:verifier-calls}
\end{table}

\begin{figure}[t]
\centering
\begin{minipage}[t]{\linewidth}
\centering
\includegraphics[width=\linewidth]{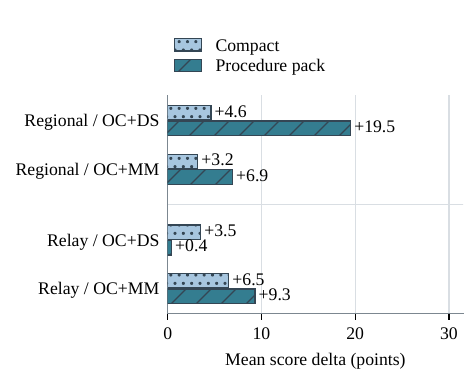}\\[-0.5ex]
\small \textbf{(A)} Procedure injection
\end{minipage}
\vspace{1.0ex}
\begin{minipage}[t]{\linewidth}
\centering
\includegraphics[width=\linewidth]{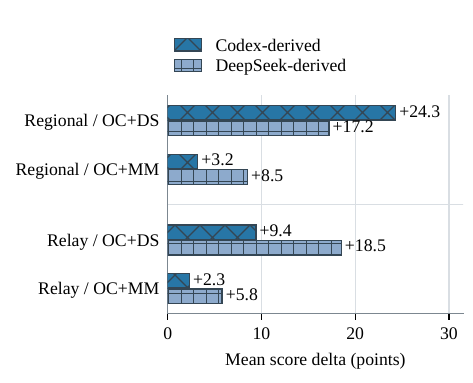}\\[-0.5ex]
\small \textbf{(B)} Memory accumulation
\end{minipage}
\caption{Workspace-support ablations for two OpenCode systems on Regional Coverage and Relay Constellation. Bars show five-case mean normalized-score deltas relative to the system- and family-specific baseline: no procedure in Panel A and no memory in Panel B. Positive values indicate higher normalized score under the support condition.}
\label{fig:ablation-effects}
\end{figure}

The second mechanism is \textbf{case-adaptive search with incumbent preservation}. Strong systems do not only apply a family-level recipe; they infer the local structure of the particular instance. In Revisit Constellation, the useful search changes once the required revisit floor is reached: additional observations no longer improve the primary metric, so the run turns to reducing satellite count while preserving a verified incumbent. In Relay Constellation, Codex CLI + GPT-5.4 searches for bottleneck bridge relays and keeps verified service plans while testing higher-risk link activations. In Regional Coverage memory transfer, useful prior experience broadens the search across regions and helps preserve the best verified candidate when later variants regress.

This case-adaptive behavior is consistent with the cases where agent rows score above solver-reference rows. The baselines are specialized and systematic, but their adaptation policy is fixed before the run. In the inspected traces, strong runs synthesize a local search policy during the run, using case files and verifier feedback to identify the current bottleneck and choose the next repair or search move. The same loop can fail when verified incumbents are not preserved or submitted.

The ablations below focus on whether workspace support improves the solution-construction side of this loop.

\subsection{Ablations}
\label{sec:ablations}

We test two prompt/workspace supports that target solution construction for OpenCode + DeepSeek V4 Pro and OpenCode + MiniMax M2.7. We choose these two systems because they share the OpenCode harness, and these two families because they place OpenCode + DeepSeek V4 Pro in a middle band, well below the stronger systems but short of total failure, where workspace support has room to register a measurable change. \textbf{Procedure injection} adds human-written task procedures to the workspace. \textbf{Memory accumulation} adds notes distilled from prior agent runs on training cases. Figure~\ref{fig:ablation-effects} reports five-case mean normalized-score deltas on Regional Coverage and Relay Constellation. The appendix reports the underlying system-level rows.

Human-written domain procedures help most when they turn construction into local verifier-guided search. For OpenCode + DeepSeek V4 Pro, the full procedure pack gives the largest Regional Coverage gain, consistent with a loop that generates candidate strips, scores marginal coverage, and preserves verified incumbents. Relay Constellation is more conjunctive: placement, routing, and scheduling must align before service appears, so procedure support gives only small gains for DeepSeek. For OpenCode + MiniMax M2.7, procedures produce smaller Regional gains and larger Relay gains from a zero-score baseline. This contrast suggests that procedures can reduce empty-service collapses, but they do not by themselves supply the coordinated search needed for high relay value.

Memory notes transfer agent-derived candidate families, verifier-backed acceptance thresholds, and repair habits. The effect is strongest for OpenCode + DeepSeek V4 Pro, especially in Regional Coverage and in Relay Constellation with DeepSeek-derived memory. OpenCode + MiniMax M2.7 also improves under memory on both families, but the gains are smaller and the resulting scores remain bounded by the system's ability to instantiate and repair the remembered plan family.

\begin{sloppypar}
Across both supports, the main pattern is conditional transfer rather than monotone improvement. Prompt/workspace support helps when it improves search discipline, incumbent preservation, or reusable repair for the active bottleneck. It does not replace task formulation, implementation calibration, or case-specific verifier evidence.
\end{sloppypar}

\section{Conclusion}
\label{sec:conclusion}

AstroAgentBench evaluates LLM agent systems on executable space mission planning tasks whose submissions are checked for feasibility and scored by mission value. Across seven task families, the results show nontrivial planning ability and clear brittleness: the best agent rows approach or exceed solver-reference scores on several families, but weaker systems often fail to produce high-value valid plans, and even strong systems lose quality on geometric, product-level, or design-heavy tasks.

The trace and ablation results point to the same pattern. High-scoring runs calibrate agent-written implementations against verifier feedback, preserve verified incumbents, and adapt search to case-specific bottlenecks. Prompt/workspace support helps when it supplies the missing part of this loop. Current agent systems can sometimes build useful case-specific solvers, but their performance depends on keeping task formulation, physical-model calibration, search, and final handoff aligned.

\section*{Limitations}

The main empirical limitation is scale. The reported evaluation covers five held-out cases for each of the seven task families, 35 cases in total, across five agent systems. Broader coverage over more cases, systems, seeds, and difficulty settings would require many more long interactive agent runs. LLM API cost is the primary constraint; wall-clock time and local compute are secondary.

The solver-reference rows are quality anchors rather than certified optima. The underlying planning families are combinatorial and include hard scheduling, coverage, routing, and design subproblems, so finding universal optima at benchmark scale is often impractical. When an agent approaches or outscores a solver row, the comparison is to the implemented reference set under the normalized scorer, not to a universal optimum.

The ablations are diagnostic rather than exhaustive. Procedure injection and memory accumulation target two plausible sources of workspace support, but they do not cover all harness designs, feedback channels, prompt structures, search tools, memory policies, or human-in-the-loop workflows. Their effects should therefore be read as evidence about the active failure layers in this setup.

The benchmark reports one sampled difficulty regime. The generators could create larger constellations, denser demand patterns, longer horizons, and higher asset counts. We do not report scaling curves across those regimes, so the results do not establish how the evaluated systems would behave as mission scale increases.

\section*{Ethical Considerations}

Space mission-planning technology has dual-use potential. Better automated planning can support scientific observation, disaster response, communications, and operations analysis, but similar capabilities could also support surveillance, military planning, or strategic space operations. AstroAgentBench is designed as an evaluation benchmark rather than an operational autonomy stack: it uses compact physical models, public or representative inputs, and offline verifiers, and it does not provide procedures for commanding spacecraft or ground systems. Extensions toward full-fidelity mission simulation or direct operational interfaces would change this risk profile.

The results also argue against direct operational reliance on current LLM agents. Several systems produce plausible artifacts that fail external checks or recover little mission value. Any use of agent-generated plans in real missions would require domain-expert review, certified planning software, independent verification, and human authority over operational decisions.

The benchmark uses generated cases, public source material, and normalized public planning instances, and its curation respects the applicable licenses for public sources. The released code and data are under the MIT License, and normalized instances remain under their original distribution terms. It does not collect personal data or human-subject annotations. Future releases should maintain these boundaries.

To improve engineering efficiency and writing clarity, we use AI-powered tools for automated code completion and language refinement. Nevertheless, all benchmark implementations, experimental scripts, and published results undergo thorough manual verification by the authors. We also perform random audits to confirm that released materials are free of sensitive information, personally identifiable data, and harmful content.

\section*{Acknowledgments}

This work is in part supported by the New Generation Artificial Intelligence-National Science and Technology Major Project (2025ZD0123502).

\bibliographystyle{plainnat}
\bibliography{references}

\clearpage
\beginappendix
\raggedbottom

\section{Benchmark Landscape}
\label{app:benchmark-landscape}

Table~\ref{tab:benchmark-landscape} compares AstroAgentBench with related benchmarks by setting, task-family breadth, and verification mechanism. The neighboring benchmarks cover symbolic or temporal planning, single-family satellite scheduling, and remote-sensing tool-use tasks.

\begin{center}
\small
\setlength{\tabcolsep}{3.4pt}
\renewcommand{\arraystretch}{1.08}
\begin{tabular}{>{\raggedright\arraybackslash}p{0.19\textwidth}>{\raggedright\arraybackslash}p{0.29\textwidth}>{\raggedright\arraybackslash}p{0.16\textwidth}>{\raggedright\arraybackslash}p{0.28\textwidth}}
\hline
\rowcolor{headerrow}
\textbf{Benchmark} & \textbf{Setting} & \textbf{Task families} & \textbf{Verification} \\
\hline
\rowcolor{solverrow}
SPOT-5 / ROADEF \citep{lemaitrea2002selecting} & Agile-satellite photograph selection & One photography-selection family & Combinatorial constraints \\
\rowcolor{solverrow}
SatNet \citep{goh2022satnet} & Ground-station contact assignment & One communication-scheduling family & Contact windows, antenna occupancy, demand metrics \\
\rowcolor{solverrow}
AEOS-Bench \citep{wang2026towards} & Realistic Earth-observation constellation scheduling & One scheduling family at large scale & Orbital dynamics, resources, scheduling metrics \\
\rowcolor{solverrow}
EOS-Bench \citep{yin2026eosbench} & Earth-observation satellite scheduling & One scheduling family with scale regimes & Orbital dynamics, platform constraints, scheduling metrics \\
\hline
\rowcolor{agentrow}
Planning benchmarks \citep{valmeekam2023planning,valmeekam2023planbench,ding2025tcp} & Symbolic action planning and temporal constraints & Multiple symbolic domains & Formal action and temporal-validity checks \\
\rowcolor{agentrow}
Remote-sensing agent benchmarks \citep{shabbir2025thinkgeo} & Tool-augmented remote-sensing analysis & Multiple remote-sensing tasks & Tool outputs and answer checks \\
\rowcolor{agentrow}
AstroAgentBench & Executable space mission planning & Seven mission-planning families & Schema, timing, geometry, resources, mission-value scoring \\
\hline
\end{tabular}
\captionof{table}{Benchmark-landscape comparison. Blue rows are agent-facing; yellow rows are solver-facing. Agent-facing means the benchmark presents tasks through prompts or a prepared workspace; verification summarizes how submitted outputs are checked.}
\label{tab:benchmark-landscape}
\end{center}
\vspace{0.75\baselineskip}

\section{Physical and Astrodynamics Models}
\label{app:physical-models}

This appendix records the physical and astrodynamics assumptions used by each task family. The benchmark does not impose a single simulator across families: some tasks use precomputed windows or abstract conflict variables, while others evaluate orbit propagation, frame conversion, access geometry, pointing limits, resource accounting, and service scoring.

For families that compute geometry from orbital state, feasibility is evaluated in a fixed physical order: propagate spacecraft state, transform inertial states into an Earth-fixed frame, then compute visibility, pointing, surface intersection, or link geometry.

\begin{table*}[!t]
\centering
\begingroup
\footnotesize
\renewcommand{\arraystretch}{1.12}
\setlength{\tabcolsep}{0pt}
\begin{tabular*}{\textwidth}{@{\extracolsep{\fill}}lccccccc@{}}
\hline
\textbf{Aspect} & \textbf{AEOSSP} & \textbf{SatNet} & \textbf{SPOT-5} & \textbf{Stereo} & \textbf{Regional} & \textbf{Revisit} & \textbf{Relay} \\
\hline
\rowcolor{black!10}
\multicolumn{8}{l}{\textit{State and reference geometry}} \\
Propagation & SGP4 (TEME) & precomputed & abstracted & SGP4 (TEME) & SGP4 (TEME) & J2 & J2 \\
Frame conversion & GCRF--ITRF & / & / & GCRF--ITRF & GCRF--ITRF & GCRF--ITRF & GCRF--ITRF \\
Surface model & WGS84 & / & / & WGS84 & WGS84 & spherical \(R_E\) & spherical \(R_E\) \\
\hline
\rowcolor{black!10}
\multicolumn{8}{l}{\textit{Geometry and action feasibility}} \\
Pointing variable & off-nadir & / & / & along/across & roll & off-nadir & / \\
Transition check & slew; settle & setup; teardown & tuple conflicts & slew; settle & roll slew & slew; settle & / \\
\hline
\rowcolor{black!10}
\multicolumn{8}{l}{\textit{Resources}} \\
Energy resource & battery & / & / & / & battery & battery & / \\
Capacity resource & / & antenna occupancy & recorder cap & / & duty limit & / & link capacity \\
\hline
\end{tabular*}
\endgroup
\caption{Physical and astrodynamics model ingredients by family. A slash marks an inactive layer because the family uses a higher-level abstraction for that part of the physical problem. SGP4 natively produces TEME states, which the propagation wrappers convert into GCRF and ITRF. Revisit and Relay convert geodetic targets or endpoints into Earth-fixed coordinates; the surface-model row names the Earth body used for altitude and blockage checks.}
\label{tab:physical-contract-inventory}
\end{table*}

Three modeling regimes recur across Table~\ref{tab:physical-contract-inventory}. SatNet and SPOT-5 are abstraction-heavy: their physical history enters the task through view periods, maintenance windows, forbidden tuples, camera domains, and capacity fields. AEOSSP, Stereo Imaging, and Regional Coverage keep the satellite set fixed and compute observation geometry from submitted actions. Revisit and Relay expose architecture variables: a solution supplies satellite states, and the model propagates those states before evaluating target visibility or communication service.

\paragraph{Propagation and Frames.} AEOSSP, Stereo Imaging, and Regional Coverage use two-line elements (TLEs): compact mean-element records for cataloged Earth satellites. Their satellite states are propagated with SGP4, the standard TLE propagator, which natively returns position and velocity in the TEME frame; the propagation libraries convert TEME states into the GCRF inertial and ITRF Earth-fixed frames before any geometry is computed. Revisit and Relay instead propagate submitted or added GCRF states with deterministic J2 dynamics, which model central-body gravity plus the dominant oblateness perturbation. For AEOSSP, Regional Coverage, Revisit, and Relay, the resulting GCRF state is transformed into ITRF/ECEF before it is compared with Earth-fixed targets, grid samples, ground stations, or relay endpoints; Stereo Imaging queries Earth-fixed states from its propagator directly.

\paragraph{Access and Pointing.} For satellite \(i\), ground object \(j\), and time \(t\), let \(\epsilon_{ij}(t)\) be the elevation angle of the satellite above the local horizon at \(j\), \(R_{ij}(t)\) the slant range, and \(\theta_{ij}(t)\) the payload pointing angle away from the allowed boresight or nadir direction. Let \(\epsilon_j^{\min}\) be the required minimum elevation at the target or ground endpoint, \(R_j^{\max}\) the maximum usable range when a range cap is modeled, and \(\theta_i^{\max}\) the payload's maximum pointing angle. A target observation is feasible only if the line of sight satisfies
\[
  \begin{aligned}
  \epsilon_{ij}(t)\ge \epsilon_j^{\min},\qquad
  R_{ij}(t)\le R_j^{\max},\\
  \theta_{ij}(t)\le \theta_i^{\max}.
  \end{aligned}
\]
Stereo imaging adds product-level constraints on overlap, convergence angle, temporal separation, and pixel-scale ratio. Regional coverage derives strip footprints from roll angle and sensor field of view, then scores weighted sample coverage.

\paragraph{Slew and Settling.} Same-satellite action sequences must leave enough time for retargeting. For a scalar angular change \(\Delta\theta\), maximum angular rate \(\omega_{\max}\), maximum angular acceleration \(\alpha_{\max}\), settling time \(t_s\), and threshold \(\theta_c=\omega_{\max}^2/\alpha_{\max}\), use the first branch below when \(\Delta\theta\le\theta_c\) and the second otherwise:
\[
  \begin{aligned}
  T_{\mathrm{slew}}(\Delta\theta)=
  \begin{cases}
    2\sqrt{\Delta\theta/\alpha_{\max}},\\
    \Delta\theta/\omega_{\max}+\omega_{\max}/\alpha_{\max}.
  \end{cases}
  \end{aligned}
\]
Consecutive actions \(a_k,a_{k+1}\) on the same satellite require
\[
  s_{k+1}-e_k\ge T_{\mathrm{slew}}(\Delta\theta_{k,k+1})+t_s.
\]

\paragraph{Energy.} Battery-constrained families integrate state of charge over model time segments. Let \(E_i(t)\) be the battery state of charge, \(E_i^{\max}\) its capacity, \(\Delta t\) a segment duration in seconds, and \(P_i^{\mathrm{chg}}(t)\) and \(P_i^{\mathrm{load}}(t)\) the charge and load powers. A generic update is
\[
  \begin{aligned}
  E_i(t+\Delta t)&=\min(E_i^{\max},\,E_i(t)+\Delta E_i(t)),\\
  \Delta E_i(t)&=\frac{(P^{\mathrm{chg}}_i(t)-P^{\mathrm{load}}_i(t))\Delta t}{3600}.
  \end{aligned}
\]
Hard feasibility requires
\[
  E_i(t)\ge 0\qquad\forall i,t.
\]
Load terms depend on the family and may include idle, imaging, and slew or maneuver power.

\paragraph{Communication and Routing.} SatNet separates antenna occupation from useful communication time. Setup and teardown consume the occupied interval \(O_k\), while only \([s_k,e_k]\) contributes service. Relay constellation separates submitted link activations from routed service. The service model builds \(G_t(S)\), allocates feasible paths, and computes latency as
\[
  \ell_{d,t}=1000\,L_{d,t}/c,
\]
where \(L_{d,t}\) is the selected path length and \(c\) is the speed of light.

\begin{table*}[!t]
\centering
\scriptsize
\setlength{\tabcolsep}{3pt}
\renewcommand{\arraystretch}{1.15}
\begin{tabular}{>{\raggedright\arraybackslash}p{0.13\textwidth}>{\raggedright\arraybackslash}p{0.24\textwidth}>{\raggedright\arraybackslash}p{0.41\textwidth}>{\raggedright\arraybackslash}p{0.16\textwidth}}
\hline
\textbf{Family} & \textbf{Submitted solution} & \textbf{Hard-validity layers} & \textbf{Native metrics} \\
\hline
AEOSSP & point-observation actions & task window; required duration; sensor type; visibility; off-nadir; same-satellite overlap; slew; battery & completion; turnaround; energy \\
SatNet & antenna-track rows & view-period containment; setup; teardown; antenna occupancy; maintenance; request/resource match; minimum duration & unsatisfied demand; satisfied requests; tracking hours \\
SPOT-5 & one camera-mode assignment per photograph & domain membership; binary conflicts; ternary conflicts; multi-orbit memory cap & profit; memory use \\
Stereo Imaging & raw observation actions & timing; access interval; solar elevation; off-nadir; overlap; slew; stereo-product geometry & target coverage; product quality \\
Regional Coverage & roll-only strip actions & time grid; duration; sensor band; strip intersection; overlap; roll slew; battery; duty limit; minimum regional coverage & weighted coverage; actions; battery \\
Revisit Constellation & initial satellite states; observation actions & satellite cap; orbit bounds; visibility; range; off-nadir; timing; overlap; slew; battery & revisit gap; satellite count \\
Relay Constellation & added satellite states; link activations & orbit bounds; ground-link geometry; inter-satellite geometry; overlap; endpoint and satellite link caps; Earth blockage & service; latency; added satellites \\
\hline
\end{tabular}
\caption{Task constraints and metrics inventory. Each row names what a valid submitted solution can contain, which hard-validity layers are applied, and which native metrics are reported for valid submissions.}
\label{tab:task-inventory}
\end{table*}

\section{Task Contracts and Metrics}
\label{app:task-contracts}

Let \(I\) denote a benchmark instance, \(S\) a submitted solution, and \(f\) a task family. The family validity indicator is
\[
  V_f(I,S)\in\{0,1\}.
\]
Here \(V_f(I,S)=1\) means that the submitted solution satisfies the hard constraints for family \(f\). Invalid or missing submissions receive score zero in aggregate tables. For any scalar \(x\), let \([x]_0^1=\min(1,\max(0,x))\). The higher-is-better score reported in cross-family tables is
\[
  \begin{aligned}
  Q_f(I,S)=
  \begin{cases}
    N_f(I,S), & V_f(I,S)=1,\\
    0, & \text{otherwise},
  \end{cases}
  \end{aligned}
\]
where \(N_f\) maps the family-native metric to the normalized scale defined below. Native metrics remain family-specific; \(Q_f\) is only the common reporting score.

Table~\ref{tab:task-inventory} lists each family's submitted object, hard-validity layers, and native metrics. The derived quantities below are computed from submitted primitive actions, states, assignments, or link intervals. Table~\ref{tab:case-splits} gives the per-family case counts behind the 116-case suite and the 35-case held-out evaluation.

\begin{table}[t]
\centering
\footnotesize
\begin{tabular}{lrrr}
\hline
\textbf{Family} & \textbf{Total} & \textbf{Train} & \textbf{Test} \\
\hline
AEOSSP & 30 & 10 & 5 \\
SatNet & 5 & 0 & 5 \\
SPOT-5 & 21 & 10 & 5 \\
Stereo & 15 & 10 & 5 \\
Regional & 15 & 10 & 5 \\
Revisit & 15 & 10 & 5 \\
Relay & 15 & 10 & 5 \\
\hline
All & 116 & 60 & 35 \\
\hline
\end{tabular}
\caption{Per-family case counts. Test is the held-out split used by the main evaluation and all ablations. AEOSSP additionally provides 15 auxiliary-split cases (test\_easy, test\_hard, test\_horizon\_2022); SatNet contributes five published instances, all held out; SPOT-5's train and test lists overlap on two instances by design, and eight further published instances are in neither split.}
\label{tab:case-splits}
\end{table}

\paragraph{AEOSSP.} Let \(\mathcal{J}\) be the task set. Task \(j\in\mathcal{J}\) has release time \(r_j\), deadline \(d_j\), required duration \(\delta_j\), required sensor type \(\sigma_j\), and weight \(w_j\). Satellite \(i\) has sensor type \(\sigma_i\). A submitted point-observation action is
\[
  a_k=(i_k,j_k,s_k,e_k),
\]
where satellite \(i_k\) observes task \(j_k\) from start time \(s_k\) to end time \(e_k\). The schedule must satisfy
\[
  \begin{aligned}
  r_{j_k}&\le s_k<e_k\le d_{j_k},\\
  e_k-s_k&=\delta_{j_k},\qquad \sigma_{i_k}=\sigma_{j_k},
  \end{aligned}
\]
as well as continuous visibility, off-nadir, same-satellite non-overlap, slew-plus-settle, and battery constraints. Let \(C(S)\subseteq\mathcal{J}\) be the tasks completed by at least one valid action, and let \(t^{\mathrm{done}}_j\) be the earliest completion time for completed task \(j\). The completed-weight fraction \(\mathrm{WCR}\) and completed-task fraction \(\mathrm{CR}\) are
\[
  \begin{aligned}
  \mathrm{WCR}(S)&=\frac{\sum_{j\in C(S)}w_j}{\sum_{j\in\mathcal{J}} w_j},\\
  \mathrm{CR}(S)&=\frac{|C(S)|}{|\mathcal{J}|}.
  \end{aligned}
\]
Turnaround time \(\mathrm{TAT}\) is the mean \(t^{\mathrm{done}}_j-r_j\) over \(j\in C(S)\), and power consumption \(\mathrm{PC}\) is gross watt-hour use over the horizon. With mission horizon \(H\) and case energy budget \(E_{\mathrm{bud}}\), the reporting score is
\[
  \begin{aligned}
  N_{\mathrm{AEOSSP}}=100(&0.45\,\mathrm{WCR}+0.20\,\mathrm{CR}\\
  &+0.20[1-\mathrm{TAT}/H]_0^1\\
  &+0.15[1-\mathrm{PC}/E_{\mathrm{bud}}]_0^1).
  \end{aligned}
\]

Here \(E_{\mathrm{bud}}\) is the sum of satellite battery capacities, used as an energy normalization constant. If no task is completed, \(\mathrm{TAT}\) is undefined and the reporting score is zero.

\emph{Worked example.} On AEOSSP test case 1, the verifier accepts the schedule from Claude Code + Claude Opus 4.6: it completes 1,415 of 1,947 tasks, with completed weight 3,669 out of 5,353. The verifier reports \(\mathrm{TAT}=1071.23\) s and \(\mathrm{PC}=18621.35\) Wh. With \(H=43200\) s and \(E_{\mathrm{bud}}=31000\) Wh, substitution gives
\[
\begin{aligned}
N_{\mathrm{AEOSSP}}
&\approx 100\Bigl(0.45\,\frac{3669}{5353}\\
&\quad+0.20\,\frac{1415}{1947}\\
&\quad+0.20\Bigl(1-\frac{1071.23}{43200}\Bigr)\\
&\quad+0.15\Bigl(1-\frac{18621.35}{31000}\Bigr)\Bigr)\\
&\approx 70.87.
\end{aligned}
\]
The five case scores are 70.87, 74.45, 73.89, 73.05, and 0.00, whose arithmetic mean is 58.45 at the reported precision. The fifth submission is parsed as an empty schedule and receives zero.

\paragraph{SatNet.} Let \(\mathcal{J}\) be the request set. Request \(j\) belongs to a subject group \(g(j)\), requests communication duration \(D_j\), and defines setup and teardown times \(\alpha_j\) and \(\beta_j\). A submitted track is
\[
  a_k=(j_k,C_k,s_k,e_k),
\]
where request \(j_k\) uses antenna or antenna-array resource \(C_k\) during transmission interval \([s_k,e_k]\). The occupied antenna interval includes setup and teardown:
\[
  O_k=[s_k-\alpha_{j_k},\,e_k+\beta_{j_k}).
\]
The transmission interval must lie inside a valid view period, all occupied intervals sharing an antenna must be disjoint, occupied intervals must avoid maintenance, the resource must be compatible with the request, and \(e_k-s_k\) must meet the per-track minimum duration rule. Let \(A_j\) be the allocated communication time for request \(j\), capped at \(D_j\). Let \(\mathcal{G}\) be the subject groups and \(\mathcal{J}_g=\{j\in\mathcal{J}:g(j)=g\}\). The unsatisfied fraction for group \(g\) is
\[
  u_g=\frac{\max\left(\sum_{j\in\mathcal{J}_g}D_j-\sum_{j\in\mathcal{J}_g}A_j,0\right)}{\sum_{j\in\mathcal{J}_g}D_j}.
\]
SatNet reports
\[
  U_{\mathrm{rms}}=\left(\frac{1}{|\mathcal{G}|}\sum_{g\in\mathcal{G}}u_g^2\right)^{1/2},\qquad
  U_{\max}=\max_{g\in\mathcal{G}}u_g,
\]
with lower values preferred. The reporting score is
\[
  \begin{aligned}
  N_{\mathrm{SatNet}}=100(&0.75[1-U_{\mathrm{rms}}]_0^1\\
  &+0.25[1-U_{\max}]_0^1).
  \end{aligned}
\]

\paragraph{SPOT-5.} Let \(\mathcal{P}\) be the candidate photograph set. Photograph \(p\) has profit \(v_p\) and allowed camera-mode domain \(D_p\), which contains only nonzero camera modes; submitting value \(0\), equivalently the all-zero assignment, denotes not selecting the photograph. Binary variable
\[
  x_{p,m}\in\{0,1\},\qquad p\in\mathcal{P},\ m\in D_p,
\]
encodes whether \(p\) is selected with mode \(m\). A valid assignment chooses at most one nonzero mode for each photograph:
\[
  \sum_{m\in D_p}x_{p,m}\le 1
\]
and must satisfy every listed binary or ternary conflict. For a forbidden tuple of arity \(r\) over photographs \(p_1,\ldots,p_r\), with forbidden modes \(m_1,\ldots,m_r\),
\[
  \sum_{\ell=1}^{r}x_{p_\ell,m_\ell}\le r-1.
\]
In multi-orbit instances, \(w_{p,m}\) is the memory consumption of selecting photograph \(p\) with mode \(m\), divided by a fixed normalizing constant and rounded, and memory use is capped by
\[
  \sum_{p\in\mathcal{P}}\sum_{m\in D_p} w_{p,m}x_{p,m}\le 200.
\]
The objective metric is selected profit,
\[
  P(S)=\sum_{p\in\mathcal{P}}v_p\sum_{m\in D_p}x_{p,m}.
\]
If \(P_{\mathrm{all}}\) is the unconstrained sum of available photograph profits, the reporting score is
\[
  N_{\mathrm{SPOT5}}=100\left[\frac{P(S)}{0.5P_{\mathrm{all}}}\right]_0^1.
\]

\paragraph{Stereo Imaging.} Let \(\mathcal{J}\) be the target set. A submitted raw observation is
\[
  a_k=(i_k,j_k,s_k,e_k,\alpha_k,\beta_k),
\]
where satellite \(i_k\) observes target \(j_k\) during \([s_k,e_k]\), and \(\alpha_k\) and \(\beta_k\) are along-track and cross-track steering angles. Let \(t_k=(s_k+e_k)/2\) be the midpoint time. Action-level validity checks timing, access interval, solar elevation, off-nadir pointing, same-satellite overlap, and slew. A pair \((a_p,a_q)\) can become a stereo product only when the two observations share a target and also satisfy the following product-level constraints. Here \(o_{pq}\) is overlap fraction, \(\gamma_{pq}\) is convergence angle, \(\eta_p\) and \(\eta_q\) are effective pixel scales, and the subscripted constants are case parameters:
\[
  \begin{aligned}
  |t_p-t_q|&\le \Delta_{\max},\\
  o_{pq}&\ge o_{\min},\\
  \gamma_{\min}&\le \gamma_{pq}\le \gamma_{\max},\\
  \frac{\max(\eta_p,\eta_q)}{\min(\eta_p,\eta_q)}&\le \eta_{\max}.
  \end{aligned}
\]
Tri-stereo products additionally require three compatible observations and a near-nadir anchor. Let \(\mathcal{P}_j(S)\) be the valid stereo or tri-stereo products covering target \(j\), and let \(Q(p)\) be product quality for product \(p\). Per-target quality is
\[
  Q_j(S)=
  \begin{cases}
    \max_{p\in\mathcal{P}_j(S)} Q(p), & \mathcal{P}_j(S)\ne\emptyset,\\
    0, & \mathcal{P}_j(S)=\emptyset,
  \end{cases}
\]
The native quality \(\bar{Q}\) is the mean of \(Q_j(S)\) over targets, and coverage is the fraction of targets with \(Q_j(S)>0\). The reporting score is \(N_{\mathrm{Stereo}}=100[\bar{Q}]_0^1\).

\paragraph{Regional Coverage.} A submitted strip action is
\[
  a_k=(i_k,s_k,\Delta_k,\rho_k),
\]
where satellite \(i_k\) images for duration \(\Delta_k\) starting at \(s_k\), and \(\rho_k\) is the signed roll angle. Validity checks time-grid alignment, duration and roll-band bounds, strip-Earth intersection, same-satellite overlap, roll slew, battery, duty limits, and optional minimum regional coverage. Let \(\mathcal{P}_{\mathrm{cov}}\) be the weighted coverage-sample set, \(w_p\) the weight of sample \(p\), and \(c_p(S)\) the number of valid strips covering \(p\). Unique weighted coverage is
\[
  \mathrm{Cov}_w(S)=
  \frac{\sum_{p\in\mathcal{P}_{\mathrm{cov}}} w_p\,\mathbf{1}\{c_p(S)\ge 1\}}{\sum_{p\in\mathcal{P}_{\mathrm{cov}}}w_p}.
\]
Let \(\mathcal{R}\) be the configured region set. For region \(r\in\mathcal{R}\), let \(\mathcal{P}_r\subseteq\mathcal{P}_{\mathrm{cov}}\) be its samples. The per-region covered fraction is
\[
  \mathrm{Cov}_r(S)=
  \frac{\sum_{p\in\mathcal{P}_r} w_p\,\mathbf{1}\{c_p(S)\ge 1\}}{\sum_{p\in\mathcal{P}_r}w_p},
\]
and the reported coverage ratio averages regions with configured weights \(\lambda_r\):
\[
  \mathrm{Cov}(S)=
  \frac{\sum_{r\in\mathcal{R}} \lambda_r\mathrm{Cov}_r(S)}{\sum_{r\in\mathcal{R}}\lambda_r}.
\]
Let \(A(S)\) be the action count, \(A_{\max}\) the action cap, \(E_{\min}\) the minimum remaining battery, and \(E_{\max}\) the representative battery capacity. The reporting score is
\[
  \begin{aligned}
  N_{\mathrm{Regional}}=100(&0.50\mathrm{Cov}_w+0.20\mathrm{Cov}\\
  &+0.15[1-A/A_{\max}]_0^1\\
  &+0.15[E_{\min}/E_{\max}]_0^1).
  \end{aligned}
\]

\paragraph{Revisit Constellation.} Let \(\mathcal{J}\) be the target set and let the mission run from \(t_0\) to \(t_1\). Target \(j\) has expected revisit period \(g^{\mathrm{req}}_j\). A solution contains \(n\) satellite initial states,
\[
  X=\{(\mathbf{r}_i(t_0),\mathbf{v}_i(t_0))\}_{i=1}^{n}
\]
and observation actions. The validity checks include satellite-count caps, orbit bounds, visibility, slant range, off-nadir pointing, timing, overlap, slew, and battery. Let \(T_j(S)=(\tau_{j,0},\ldots,\tau_{j,L_j})\) be the ordered sequence containing \(t_0\), the successful observation midpoint times for target \(j\), and \(t_1\). The maximum revisit gap for target \(j\) is
\[
  g_j(S)=\max_{0\le \ell<L_j}(\tau_{j,\ell+1}-\tau_{j,\ell}).
\]
The reported primary metric is the mean requirement-floored gap,
\[
  G(S)=\frac{1}{|\mathcal{J}|}\sum_{j\in\mathcal{J}}\max(g_j(S),\,g^{\mathrm{req}}_j),
\]
with lower values preferred, followed by satellite count. Let \(H=t_1-t_0\). For a scalar gap \(g\) and requirement \(g_{\mathrm{req}}\), the target-level reporting curve is
\[
  \phi_{\mathrm{gap}}(g)=
  \begin{cases}
    1, & g\le g_{\mathrm{req}},\\
    0, & g\ge H,\\
    \left(\frac{H-g}{H-g_{\mathrm{req}}}\right)^2, & \text{otherwise}.
  \end{cases}
\]
The gap component averages this curve over targets:
\[
  q_{\mathrm{gap}}=\frac{1}{|\mathcal{J}|}\sum_{j\in\mathcal{J}}\phi_{\mathrm{gap}}(g_j(S)).
\]
Let \(n_{\min}\) and \(n_{\max}\) be the configured lower and upper satellite-count anchors. The scarcity bonus is
\[
  q_{\mathrm{sat}}=\left[(n_{\max}-n)/(n_{\max}-n_{\min})\right]_0^1,
\]
and it is applied only after every target reaches its revisit requirement:
\[
  \begin{aligned}
  N_{\mathrm{Revisit}}&=
    \begin{cases}
      70q_{\mathrm{gap}}, & q_{\mathrm{gap}}<1,\\
      70+30q_{\mathrm{sat}}^2, & q_{\mathrm{gap}}=1.
    \end{cases}
  \end{aligned}
\]

\paragraph{Relay Constellation.} Let \(\mathcal{D}\) be the demand set. Demand \(d\) has a weight \(\omega_d\), two endpoints, and requested sample times \(\mathcal{T}_d\). A solution contains added satellite states and link-activation intervals. At sample time \(t\), the active validated links define a graph
\[
  G_t(S)=(\mathcal{N},E_t(S)),
\]
where \(\mathcal{N}\) contains endpoints and satellites, and \(E_t(S)\) contains validated ground or inter-satellite links. Demand \(d\) is served at time \(t\) if its endpoints are connected in \(G_t(S)\) under the route-allocation rules. The per-demand service fraction is
\[
  \mathrm{service}_d(S)=
  \frac{|\{t\in\mathcal{T}_d:\ d\text{ is served at }t\}|}{|\mathcal{T}_d|}.
\]
The aggregate service fraction \(s\) is the demand-weighted mean,
\[
  s=\frac{\sum_{d\in\mathcal{D}}\omega_d\,\mathrm{service}_d(S)}{\sum_{d\in\mathcal{D}}\omega_d},
\]
and \(s_{\min}=\min_{d\in\mathcal{D}}\mathrm{service}_d(S)\). The remaining native metrics are the number of added satellites \(n\), mean latency \(\bar{\ell}\), and 95th-percentile latency \(\ell_{95}\), computed over served samples. The service core is
\[
  q_{\mathrm{svc}}=0.75s+0.25s_{\min}.
\]
If \(s<1\), then \(N_{\mathrm{Relay}}=70q_{\mathrm{svc}}\). For \(s=1\), use added-satellite anchors \(n_{\min},n_{\max}\) and latency cap \(L_{\max}\):
\[
  \begin{aligned}
  u_n&=\left[\frac{n_{\max}-n}{n_{\max}-n_{\min}}\right]_0^1,
  \qquad q_{\mathrm{sat}}=u_n^2,\\
  q_{\mathrm{lat}}&=0.60\left[1-\frac{\bar{\ell}}{L_{\max}}\right]_0^1\\
  &\quad+0.40\left[1-\frac{\ell_{95}}{L_{\max}}\right]_0^1,\\
  N_{\mathrm{Relay}}&=70+30(0.75q_{\mathrm{sat}}+0.25q_{\mathrm{lat}}).
  \end{aligned}
\]

\begin{table*}[!t]
\centering
\scriptsize
\setlength{\tabcolsep}{3pt}
\renewcommand{\arraystretch}{1.12}
\begin{tabular}{>{\raggedright\arraybackslash}p{0.13\textwidth}>{\raggedright\arraybackslash}p{0.21\textwidth}>{\raggedright\arraybackslash}p{0.58\textwidth}}
\hline
\textbf{Family} & \textbf{Reference} & \textbf{Core abstraction} \\
\hline
AEOSSP & MWIS conflict graph \citep{eddy2021maximum} & Enumerates feasible observation candidates, links incompatible candidates in a conflict graph, and selects a weighted independent set before verifier-facing repair. \\
AEOSSP & Greedy LNS \citep{antuori2025solving} & Builds a satellite-local schedule by greedy candidate insertion, then reinserts bounded neighborhoods to improve completion and turnaround. \\
SatNet & Delta-MILP \citep{claudet2021delta} & Uses a mixed-integer contact-assignment model to allocate antenna time while controlling request-level unsatisfied demand. \\
SatNet & PPO \citep{goh2022satnet} & Uses a trained reinforcement-learning policy to choose contact assignments under the SatNet request-service metric. \\
SPOT-5 & Reference lookup \citep{lemaitrea2002selecting} & Matches known held-out instances to archived challenge solutions and recomputes their profit under the benchmark verifier. \\
Stereo Imaging & CP/local search \citep{lemaitrea2002selecting} & Constructs a library of pair and tri-stereo products, then inserts and repairs products as coupled scheduling objects. \\
Stereo Imaging & Pruned MILP \citep{kim2020task} & Prunes observation windows and stereo products before solving a coverage-first mixed-integer selection model. \\
Regional Coverage & CP local search \citep{antuori2025solving} & Generates roll-only strip candidates, scores marginal weighted coverage, and repairs sequence neighborhoods with bounded CP. \\
Regional Coverage & CELF selection \citep{leskovec2007cost,nemhauser1978analysis,khuller1999budgeted} & Treats fixed strip candidates as a submodular coverage set and applies lazy marginal-gain selection with schedule filtering. \\
Revisit Constellation & J2 RGT set cover \citep{lee2024satellite} & Searches J2 repeat-ground-track shells, expands RAAN candidates, and schedules confirmed target assignments to reduce revisit gaps. \\
Revisit Constellation & RGT/APC constructive \citep{zhang2018leo,lee2020apc,mercado2025scheduling} & Builds repeat-ground-track access profiles, selects gap-improving satellites, and schedules observations by target freshness. \\
Relay Constellation & MCLP+TEG \citep{williams2026optimal,gerard2026contact} & Selects relay candidates by contact opportunity coverage, then emits route-aware link activations over a time-expanded graph. \\
Relay Constellation & UMCF/SRR \citep{grislain2022rethinking,lamothe2023dynamic} & Builds dynamic communication graphs, solves path-restricted flow relaxations, and rounds paths into verifier-compatible link intervals. \\
\hline
\end{tabular}
\caption{Solver-reference inventory for Table~\ref{tab:main-results}. The table names the public reference rows and states the core algorithmic idea for each row; evidence status and score mapping are described in the surrounding text.}
\label{tab:solver-baseline-inventory}
\end{table*}

\section{Solver Baselines}
\label{app:solver-baselines}

The solver-reference rows in Table~\ref{tab:main-results} are casewise quality anchors under the same normalization as the agent rows. For each held-out case, we score every available reference with the family rule in Appendix~\ref{app:task-contracts}. The first solver row averages the best reference per case; when a second reference exists, the second row averages the second-best reference. A row may therefore combine different named references across cases. SPOT-5 has one solver row because each held-out case uses one archived reference-solution source.

Five families use implemented references that write the same submitted artifact type as agents and are checked by the family verifier. Table~\ref{tab:solver-baseline-inventory} lists these public rows and their core abstractions. SatNet uses literature-reported Delta-MILP and PPO values for the same oversubscribed weeks \citep{claudet2021delta,goh2022satnet}. SPOT-5 matches held-out instances to archived challenge solutions and recomputes their profit and resource weight before normalization.

\paragraph{Score Relationship.} Appendix~\ref{app:task-contracts} defines the family score \(N_f\). Higher-is-better native metrics map directly to higher normalized scores, lower-is-better metrics are inverted, and Revisit and Relay keep their gates: satellite-count or latency terms enter only after the revisit or service requirement is met.

\section{Agent Environment}
\label{app:agent-workspace}

This appendix records the run configuration used for the agent rows in Table~\ref{tab:main-results}. An evaluated system consists of a configured model, a harness, and the prepared workspace in which the run occurs. The main-experiment workspace supplies the task brief, case files, output contract, agent-local instructions, and verifier helper before the harness is launched. Table~\ref{tab:workspace-environment} summarizes the layers exposed to the run.

\Needspace{13\baselineskip}
The main-experiment workspace is initialized with the following structure:
\begin{Verbatim}[fontsize=\scriptsize, frame=single, framesep=1mm]
workspace/
  .agents/skills/ (or .claude/skills)
  case/
  AGENTS.md (or CLAUDE.md)
  README.md
  verifier
\end{Verbatim}

The official scorer consumes the final artifact after the run stops. The workspace always includes a runnable verifier helper as an opaque binary; ablation runs may add procedure or memory materials, whose condition definitions appear with the corresponding ablation rows in Appendix~\ref{app:ablation-results}.

\begin{table*}[!t]
\centering
\footnotesize
\setlength{\tabcolsep}{4pt}
\begin{tabular}{>{\raggedright\arraybackslash}p{0.18\textwidth}>{\raggedright\arraybackslash}p{0.74\textwidth}}
\hline
\textbf{Layer} & \textbf{Contents exposed to the run} \\
\hline
Task package & Rendered family brief, rendered task prompt, case files, required output contract, and the runnable opaque verifier helper. \\
Runtime tools & Shared Linux container with Python 3.13, Node.js 24.x, OpenJDK 17, shell tools, Git, JSON utilities, file-search utilities, and the evaluated agent CLI packages. \\
Python libraries & Pinned astrodynamics, geometry, optimization, and data libraries, including Brahe, Basilisk, Orekit bindings, Skyfield, OR-Tools, PuLP, NetworkX, NumPy, SciPy, pandas, Shapely, pyproj, and plotting utilities. \\
Agent-local guidance & The Brahe skill document is installed under \texttt{.agents/skills/} for each harness where supported. Procedure-injection and memory-accumulation experiments add only the configured procedures or prior-run notes for the corresponding condition. \\
Execution controls & A fixed two-hour wall-clock limit, 8 CPU allocation, 32 GB memory limit, 16 GB shared-memory allocation, no human repair during the run, and post-run official scoring of the final submitted artifact. \\
Collected evidence & Final submitted artifact, official verifier output, aggregate metrics, and the harness-specific session logs needed for later trace analysis. \\
\hline
\end{tabular}
\caption{Prepared run environment. The benchmark controls the task package and official scorer; the evaluated system controls the interaction policy inside the prepared workspace.}
\label{tab:workspace-environment}
\end{table*}

Table~\ref{tab:agent-system-versions} reports the harness package version and reasoning configuration for the five evaluated systems, and Table~\ref{tab:runtime-library-versions} records the pinned Python libraries in the base runtime. The evaluated-system names carry the model identities.

\begin{table*}[!t]
\centering
\begingroup
\footnotesize
\setlength{\tabcolsep}{5pt}
\begin{tabular}{>{\raggedright\arraybackslash}p{0.34\textwidth}>{\raggedright\arraybackslash}p{0.34\textwidth}>{\raggedright\arraybackslash}p{0.20\textwidth}}
\hline
\textbf{Evaluated system} & \textbf{Harness package} & \textbf{Reasoning configuration} \\
\hline
Claude Code + Claude Opus 4.6 & \texttt{@anthropic-ai/claude-code} 2.1.123 & high \\
Codex CLI + GPT-5.4 & \texttt{@openai/codex} 0.125.0 & high \\
Kimi CLI + Kimi K2.6 & \texttt{kimi-cli} 1.40.0 & thinking \\
OpenCode + MiniMax M2.7 & \texttt{opencode-ai} 1.14.30 & thinking \\
OpenCode + DeepSeek V4 Pro & \texttt{opencode-ai} 1.14.30 & max \\
\hline
\end{tabular}
\caption{Harness package and reasoning configuration for the five evaluated systems. Versions are taken from the shared base runtime.}
\label{tab:agent-system-versions}
\endgroup

\vspace{0.75\baselineskip}

\begingroup
\small
\setlength{\tabcolsep}{4pt}
\begin{tabular}{>{\raggedright\arraybackslash}p{0.20\textwidth}>{\raggedright\arraybackslash}p{0.72\textwidth}}
\hline
\textbf{Layer} & \textbf{Pinned versions in the base runtime} \\
\hline
Astrodynamics & \texttt{bsk==2.9.0}; \texttt{brahe==1.4.2}; \texttt{orekit-jpype==13.1.4.0}; \texttt{shapely==2.1.2}; \texttt{skyfield==1.54}; \texttt{pyproj==3.7.2}; \texttt{pygmo==2.19.8}. \\
Optimization & \texttt{ortools==9.15.6755}; \texttt{pulp==3.3.0}; \texttt{networkx==3.6.1}; \texttt{numpy==2.2.6}; \texttt{scipy==1.17.1}; \texttt{sympy==1.14.0}. \\
Scientific & \texttt{pandas==2.3.3}; \texttt{pydantic==2.13.3}; \texttt{matplotlib==3.10.7}; \texttt{plotly==6.7.0}; \texttt{scikit-learn==1.8.0}; \texttt{datasets==4.8.4}; \texttt{huggingface-hub==1.12.0}; \texttt{kagglehub==1.0.0}; \texttt{pyyaml==6.0.3}; \texttt{pytest==9.0.3}; \texttt{tqdm==4.67.1}; \texttt{pyinstaller==6.20.0}. \\
\hline
\end{tabular}
\caption{Base runtime Python library versions available to all evaluated systems inside the shared container.}
\label{tab:runtime-library-versions}
\endgroup
\end{table*}

\Needspace{5\baselineskip}
Representative fragments from the agent-facing task material are shown in Appendix~\ref{app:prompt-fragments}. The full materials are longer because each family also includes field-level schemas and modeling details.

Agents may iterate locally during the run, but the recorded artifact is the submitted solution file present at termination. Missing files, invalid files, and valid low-quality plans are separated by the protocol. Results are interpreted at the agent-system level: model, harness, workspace, and submitted artifact together define the measured behavior.

\clearpage
\section{Representative Prompt Fragments}
\label{app:prompt-fragments}
\enlargethispage{12\baselineskip}

This appendix records selected fragments from the agent-facing task material, chosen for transparency about the contract seen by agents. The complete materials also include schemas, file descriptions, examples, and family-specific details, which are too long to reproduce here. The excerpts below preserve operative wording; line breaks may be wrapped for print, ellipses mark omitted neighboring text, and workspace-relative tokens are retained when they are part of the prompt contract.\par\nobreak\smallskip\noindent\begin{minipage}{\columnwidth}
\paragraph{Shared Workspace Rules (AGENTS.md).}\mbox{}\par
\begin{promptblockinnercompact}
...
- Treat `solution.json` as the required final deliverable unless the workspace
  says otherwise.
- If the workspace exposes a verifier helper, use it for local iteration when
  helpful.
- The run may be stopped after 2 hours. As soon as you have any valid or
  likely-valid answer, write it to `solution.json` and keep that file valid
  while you continue improving it.
- Prefer incremental improvement: preserve the best working `solution.json`
  you have, and only replace it after the replacement is written completely
  and is at least as likely to verify.
...
\end{promptblockinnercompact}
\end{minipage}\par\smallskip

\Needspace{31\baselineskip}
\promptheadingcompact{Brahe Skill.}
\begin{promptblockcompact}
---
name: brahe
description: |
  ...
---

# Brahe Skill

Curated documentation and runnable examples for the Brahe Python library.

...

## Module Map

See more examples and documents on how to use brahe:

| Topic | Reference |
|-------|-----------|
| **Time & EOP** | [docs/learn/time/index.md](docs/learn/time/index.md),
  [docs/learn/eop/index.md](docs/learn/eop/index.md) |
| **Coordinates** | [docs/learn/coordinates/index.md](docs/learn/coordinates/index.md) |
| **Propagation** | [docs/learn/orbit_propagation/index.md](docs/learn/orbit_propagation/index.md),
  [docs/learn/orbit_propagation/numerical_propagation/index.md](docs/learn/orbit_propagation/numerical_propagation/index.md) |
| **Access** | [docs/learn/access_computation/index.md](docs/learn/access_computation/index.md) |
...
\end{promptblockcompact}

\promptheadingcompact{Short Task Instructions.}
\begin{promptblockcompact}
Stereo Imaging:
Please solve the prepared stereo-imaging case using the files in `case/`.
...
Write `solution.json` at the workspace root. You have a 2-hour timeout, so
preserve a valid or likely-valid `solution.json` as soon as possible and keep
improving it in place. Prioritize normalized stereo quality first, then
improve valid stereo coverage where you can.

Relay Constellation:
Please solve the prepared relay-network augmentation case using the files in
`case/`.
...
Write `solution.json` at the workspace root. You have a 2-hour timeout, so
preserve a valid or likely-valid `solution.json` as soon as possible and keep
improving it in place. Build a feasible augmentation and link plan first, then
push service up and latency down.

Revisit Constellation:
Please solve the prepared revisit-driven constellation planning case using the
files in `case/`.
...
Write `solution.json` at the workspace root. You have a 2-hour timeout, so
preserve a valid or likely-valid `solution.json` as soon as possible and keep
improving it in place. Produce a feasible constellation and observation
schedule, then push the revisit gaps down as far as you can.
...
\end{promptblockcompact}

\Needspace{9\baselineskip}
\promptheadingcompact{AEOSSP.}
\begin{promptblockcompact}
...
Do not submit visibility claims, maneuver windows, power traces, or completion
claims.
Those are derived during validation.
...
\end{promptblockcompact}

\Needspace{16\baselineskip}
\promptheadingcompact{Stereo Imaging.}
\begin{promptblockcompact}
The output should schedule raw observations only. Stereo pairing,
tri-stereo grouping, overlap checks, convergence checks, and quality scoring
are derived during validation.
...
The satellite set is fixed by TLEs in `case/satellites.yaml`; you choose only
raw observation windows and two boresight steering angles.
`off_nadir_along_deg` tilts along the flight direction and
`off_nadir_across_deg` tilts cross-track in the satellite local frame.
...
Target access is derived from the target center at `longitude_deg`,
`latitude_deg`, and `elevation_ref_m`; it is not a submitted claim.
...
\end{promptblockcompact}

\promptheadingcompact{Regional Coverage.}
\begin{promptblockcompact}
Your job is to produce `solution.json`, a schedule of `strip_observation`
actions that maximizes unique weighted regional coverage while remaining valid.
...
Do not submit your own strip polygons, coverage claims, or access-window
identifiers. Strip geometry and coverage are derived during validation.
...
the only attitude command is signed `roll_deg`, the cross-track off-nadir
angle of the strip center. Positive and negative signs look to opposite sides
of the ground track.
...
\end{promptblockcompact}

\promptheadingcompact{Revisit Constellation.}
\begin{promptblockcompact}
- a proposed constellation at mission start
- a schedule of observation actions for that constellation
...
The goal is to keep revisit gaps small across the targets while respecting the
orbit, visibility, timing, slew, and power constraints.
...
In practical terms, feasibility comes first. After that, the solution should
drive revisit gaps down as much as possible, and efficient constellation size
matters once the required revisit quality is achieved.
...
\end{promptblockcompact}

\promptheadingcompact{Relay Constellation.}
\begin{promptblockcompact}
The modeled decision is: add a limited number of relay satellites and choose
when physical links are active. You do not submit routes, per-demand service
assignments, or latency calculations. Instead, the validator builds a
time-varying communication graph from your active links and then computes how
much demand can actually be served through that graph.
...
With multiple active demands, allocation maximizes total served demand
`weight`, then minimizes total latency, then uses deterministic path ordering
as a tie-breaker.
...
\end{promptblockcompact}

\Needspace{12\baselineskip}
\promptheadingcompact{Injected Procedures.}
\emph{Regional coverage procedure.}
\begin{promptblockcompact}
| Valid nonzero but weak | Build a small candidate table and add one
fresh-coverage strip at a time. |
...
When comparing candidate moves, use this order after each verifier run:
1. validity
2. higher `weighted_coverage_ratio`
...
\end{promptblockcompact}

\emph{Relay constellation procedure.}
\begin{promptblockcompact}
If `valid=true` and `service_fraction=0`, the next move is not latency
optimization. The next move is to complete one served path.
...
Private propagation or visibility code is a candidate generator, not proof. If
the verifier disagrees, change the candidate.
...
\end{promptblockcompact}

\promptheadingcompact{SatNet.}
\begin{promptblockcompact}
Prioritize request satisfaction and mission-level fairness: minimize `U_rms`
first, minimize `U_max` next, then use satisfied request count and valid
communication time as tie-breakers.
...
U_rms = sqrt(mean(U_i^2))
U_max = max(U_i)
...
\end{promptblockcompact}

\promptheadingcompact{SPOT-5.}
\begin{promptblockcompact}
This problem is modeled as a constrained assignment-and-selection problem
rather than as orbital propagation.
...
Assignment `0` rejects the photograph. A nonzero assignment selects it and
must be in that variable's domain. Values `1`, `2`, and `3` are mono-camera
choices. Value `13` is the dual-camera choice and is legal only when the
variable's domain explicitly contains `13`.
...
Header mismatches in `claimed_profit` or `claimed_weight` may produce warnings
rather than immediate invalidity, but the assignment vector itself must still
satisfy all domain, conflict, and capacity rules.
...
computed_profit = sum(profit_i for i where assignments[i] != 0)
computed_weight = sum(normalized_weight_i for selected variables)
computed_weight <= 200
...
\end{promptblockcompact}

\section{Uncertainty in the Main Results}
\label{app:score-uncertainty}

Table~\ref{tab:score-uncertainty} reports 95\% percentile case-bootstrap confidence intervals for the agent-system mean scores. For each family, we enumerate all \(5^5=3125\) ordered samples of five cases drawn with replacement, using the same case indices across systems. We recompute the mean for each sample and take the 2.5th and 97.5th percentiles, with linear interpolation, as the interval endpoints. All original zero-score outcomes are retained. The recorded run for each system--case pair is held fixed, so these intervals summarize case-resampling variation and do not estimate variability across repeated agent runs. With only five cases per family, they provide a limited uncertainty summary; an all-zero interval reflects five observed zero scores and does not establish zero performance on unseen cases.

\begin{table*}[!t]
\centering
\small
\setlength{\tabcolsep}{5pt}
\renewcommand{\arraystretch}{1.15}
\begin{tabular}{lccccc}
\hline
\rowcolor{headerrow}
Family & Claude & Codex & Kimi & OC+MM & OC+DS \\
\hline
AEOSSP & \shortstack{58.45\\{\scriptsize [29.06, 73.95]}} & \shortstack{72.91\\{\scriptsize [69.87, 76.15]}} & \shortstack{72.03\\{\scriptsize [69.43, 74.14]}} & \shortstack{36.44\\{\scriptsize [16.60, 52.71]}} & \shortstack{71.44\\{\scriptsize [69.15, 73.73]}} \\
Regional & \shortstack{21.22\\{\scriptsize [20.71, 22.22]}} & \shortstack{72.08\\{\scriptsize [60.04, 80.37]}} & \shortstack{73.15\\{\scriptsize [67.40, 77.38]}} & \shortstack{17.73\\{\scriptsize [9.98, 24.82]}} & \shortstack{35.79\\{\scriptsize [24.15, 44.92]}} \\
Relay & \shortstack{12.25\\{\scriptsize [0.00, 36.75]}} & \shortstack{64.91\\{\scriptsize [55.81, 77.71]}} & \shortstack{57.77\\{\scriptsize [42.85, 72.63]}} & \shortstack{0.00\\{\scriptsize [0.00, 0.00]}} & \shortstack{35.67\\{\scriptsize [11.55, 59.78]}} \\
Revisit & \shortstack{76.50\\{\scriptsize [73.72, 78.85]}} & \shortstack{69.58\\{\scriptsize [61.90, 76.01]}} & \shortstack{73.36\\{\scriptsize [69.78, 78.13]}} & \shortstack{0.00\\{\scriptsize [0.00, 0.00]}} & \shortstack{70.95\\{\scriptsize [69.98, 71.93]}} \\
SatNet & \shortstack{60.36\\{\scriptsize [29.25, 79.41]}} & \shortstack{69.45\\{\scriptsize [61.78, 77.13]}} & \shortstack{66.14\\{\scriptsize [58.88, 74.96]}} & \shortstack{37.54\\{\scriptsize [16.55, 55.85]}} & \shortstack{60.23\\{\scriptsize [50.07, 70.39]}} \\
SPOT-5 & \shortstack{43.96\\{\scriptsize [11.09, 76.83]}} & \shortstack{61.92\\{\scriptsize [44.58, 82.18]}} & \shortstack{61.92\\{\scriptsize [44.58, 82.18]}} & \shortstack{37.47\\{\scriptsize [7.15, 71.09]}} & \shortstack{60.91\\{\scriptsize [43.99, 82.17]}} \\
Stereo & \shortstack{19.16\\{\scriptsize [0.31, 55.67]}} & \shortstack{68.91\\{\scriptsize [33.64, 93.67]}} & \shortstack{0.25\\{\scriptsize [0.00, 0.75]}} & \shortstack{0.00\\{\scriptsize [0.00, 0.00]}} & \shortstack{13.00\\{\scriptsize [0.00, 38.99]}} \\

\hline
\end{tabular}
\caption{Mean normalized scores with 95\% case-bootstrap confidence intervals beneath them, computed from five held-out cases per family. Systems and model versions are those in Table~\ref{tab:main-results}; OC+MM and OC+DS denote OpenCode with MiniMax and DeepSeek, respectively. Intervals describe each system's mean and are not pairwise significance tests.}
\label{tab:score-uncertainty}
\end{table*}

\section{Supplementary Self-Hosted System Evaluation}
\label{app:qwen-supplement}

We evaluated OpenCode 1.14.19 with self-hosted Qwen3.6-27B through an OpenAI-compatible endpoint on the five held-out SPOT-5 cases in three batches. Each agent workspace was configured with 8 CPUs, 32 GB memory, and a 7,200-second run budget. Table~\ref{tab:qwen-supplement} reports the normalized scores under the same validity and scoring rules as the main evaluation. Eight of the 15 recorded attempts produced verifier-valid submissions; six produced no submission and one produced an invalid submission. This supplement is limited to the tested system configuration and SPOT-5 cases.

\begin{table*}[!t]
\centering
\small
\setlength{\tabcolsep}{6pt}
\begin{tabular}{crrrrrcr}
\hline
\rowcolor{headerrow}
Repetition & Case 8 & Case 28 & Case 1021 & Case 1403 & Case 1506 & Valid & Mean \\
\hline
1 & 100.00 & 0.00 & 54.80 & \(0.00^{\dagger}\) & 64.36 & 3/5 & 43.83 \\
2 & 100.00 & 34.37 & 0.00 & 0.00 & 0.00 & 2/5 & 26.87 \\
3 & 100.00 & 34.37 & 0.00 & 35.60 & 0.00 & 3/5 & 33.99 \\
\hline
\end{tabular}
\caption{Supplementary SPOT-5 normalized scores for OpenCode with self-hosted Qwen3.6-27B. Missing or invalid submissions receive zero, and each mean includes all five cases. \(\dagger\) The recorded attempt failed during provider/model lookup before inference and produced no submission; its zero is retained under the missing-submission rule.}
\label{tab:qwen-supplement}
\end{table*}

\clearpage
\section{Family-Native Metrics and Ablations}
\label{app:ablation-results}

This appendix reports the family-native metrics behind Table~\ref{tab:main-results} and the per-system ablation rows behind Section~\ref{sec:ablations}. The first subsection gives the raw mission metrics the source formulations optimize, so the normalized mapping in Appendix~\ref{app:task-contracts} can be read against the quantities it aggregates. The procedure-injection and memory-accumulation subsections then report five-case rows the main body only summarizes, with missing or invalid submissions scored zero: procedure injection tests whether written task procedures change search behavior, and memory accumulation tests whether prior-run notes transfer. Each ablation covers the systems and families needed for its intervention.

\subsection{Family-Native Metrics --- Main Experiment}

\begin{table*}[!t]
\centering
\small
\setlength{\tabcolsep}{6pt}
\renewcommand{\arraystretch}{1.15}
\begin{tabularx}{\textwidth}{MAAAAARR}
\hline
\multicolumn{1}{>{\columncolor{metriclabelcell}}l}{} & \multicolumn{5}{>{\columncolor{agentmetriccell}}c}{Agent systems} & \multicolumn{2}{>{\columncolor{referencemetriccell}}c}{Reference} \\
\multicolumn{1}{>{\columncolor{metriclabelcell}}l}{Family / metric} & \multicolumn{1}{>{\columncolor{agentmetriccell}}r}{Claude} & \multicolumn{1}{>{\columncolor{agentmetriccell}}r}{Codex} & \multicolumn{1}{>{\columncolor{agentmetriccell}}r}{Kimi} & \multicolumn{1}{>{\columncolor{agentmetriccell}}r}{OC+MM} & \multicolumn{1}{>{\columncolor{agentmetriccell}}r}{OC+DS} & \multicolumn{1}{>{\columncolor{referencemetriccell}}r}{Ref1} & \multicolumn{1}{>{\columncolor{referencemetriccell}}r}{Ref2} \\
\hline
\multicolumn{8}{>{\columncolor{headerrow}}l}{\textbf{AEOSSP} \,(weighted)} \\
\quad valid (/5) & 5 & 5 & 5 & 5 & 5 & 5 & 5 \\
\quad WCR (.45)\,$\uparrow$ & 0.570 & 0.708 & 0.692 & 0.201 & 0.682 & \bestref{0.758} & 0.682 \\
\quad CR (.20)\,$\uparrow$ & 0.594 & 0.736 & 0.720 & 0.176 & 0.710 & \bestref{0.790} & 0.721 \\
\quad TAT (s) (.20)\,$\downarrow$ & 9477 & 1025 & \bestagent{973} & 9369 & 987 & 1018 & 1128 \\
\quad PC (kWh) (.15)\,$\downarrow$ & 16.5 & 19.1 & 18.8 & \bestagent{9.8} & 18.7 & 19.8 & 18.5 \\
\hline
\multicolumn{8}{>{\columncolor{headerrow}}l}{\textbf{SatNet} \,(weighted)} \\
\quad valid (/5) & 4 & 5 & 5 & 4 & 5 & 5 & 5 \\
\quad $U_{\mathrm{rms}}$ (.75)\,$\downarrow$ & 0.354 & \bestagent{0.234} & 0.248 & 0.550 & 0.268 & 0.302 & 0.316 \\
\quad $U_{\max}$ (.25)\,$\downarrow$ & 0.525 & \bestagent{0.519} & 0.611 & 0.849 & 0.787 & 0.673 & 0.772 \\
\hline
\multicolumn{8}{>{\columncolor{headerrow}}l}{\textbf{Regional Coverage} \,(weighted)} \\
\quad valid (/5) & 5 & 5 & 5 & 5 & 5 & 5 & 5 \\
\quad $\mathrm{Cov}_w$ (.50)\,$\uparrow$ & 0.000 & 0.880 & 0.944 & 0.044 & 0.397 & \bestref{0.995} & 0.968 \\
\quad Cov (.20)\,$\uparrow$ & 0.000 & 0.881 & 0.945 & 0.039 & 0.374 & \bestref{0.995} & 0.970 \\
\quad actions (.15)\,$\downarrow$\,$\dagger$ & \bestagent{0.0} & 45.8 & 60.4 & 27.6 & 54.2 & --- & --- \\
\quad min batt (Wh) (.15)\,$\uparrow$\,$\dagger$ & \bestagent{495} & 494 & 494 & \bestagent{495} & 492 & \bestref{495} & 494 \\
\hline
\multicolumn{8}{>{\columncolor{headerrow}}l}{\textbf{Revisit Constellation} \,(gated)} \\
\quad valid (/5) & 5 & 5 & 5 & 0 & 5 & 5 & 5 \\
\quad gap (h)\,$\downarrow$ (gate) & \bestagent{6.80} & 7.67 & 6.84 & 48.00 & 6.84 & \bestref{6.80} & 8.83 \\
\quad sats\,$\downarrow$ (bonus)\,$\dagger$ & \bestagent{11.4} & 16.0 & 14.4 & --- & 17.0 & 17.2 & 19.8 \\
\hline
\multicolumn{8}{>{\columncolor{headerrow}}l}{\textbf{Relay Constellation} \,(gated)} \\
\quad valid (/5) & 1 & 5 & 5 & 4 & 5 & 5 & 5 \\
\quad service\,$\uparrow$ & 0.189 & \bestagent{0.935} & 0.868 & 0.000 & 0.559 & 0.933 & 0.928 \\
\quad worst-grp\,$\uparrow$ & 0.133 & \bestagent{0.684} & 0.522 & 0.000 & 0.360 & 0.633 & 0.643 \\
\hline
\multicolumn{8}{>{\columncolor{headerrow}}l}{\textbf{SPOT-5} \,(single objective)} \\
\quad valid (/5) & 3 & 5 & 5 & 4 & 5 & 5 & --- \\
\quad profit (k)\,$\uparrow$ & 68.9 & \bestagent{115.3} & \bestagent{115.3} & 45.5 & 112.1 & 112.3 & --- \\
\quad selected\,$\ddagger$ & 125.2 & 178.2 & 177.8 & 62.4 & 173.4 & --- & --- \\
\hline
\multicolumn{8}{>{\columncolor{headerrow}}l}{\textbf{Stereo Imaging} \,(single objective)} \\
\quad valid (/5) & 3 & 5 & 5 & 1 & 4 & 5 & 5 \\
\quad quality\,$\uparrow$ & 0.192 & 0.689 & 0.003 & 0.000 & 0.130 & \bestref{0.961} & 0.920 \\
\quad coverage\,$\ddagger$ & 0.197 & 0.723 & 0.003 & 0.000 & 0.188 & 0.971 & 0.944 \\
\hline
\end{tabularx}
\caption{Family-native metrics behind Table~\ref{tab:main-results}, in raw units, for the five agent systems (Claude = Claude Code + Claude Opus 4.6; Codex = Codex CLI + GPT-5.4; Kimi = Kimi CLI + Kimi K2.6; OC+MM = OpenCode + MiniMax M2.7; OC+DS = OpenCode + DeepSeek V4 Pro) and the two strongest reference solvers per family. Pale green, blue, and yellow column shading separate metric labels, agent-system columns, and reference columns; darker bold cells mark the best value in rows with a displayed preference direction, with ties all marked. Each row is one native term the family score uses; the parenthetical weight is its coefficient in the normalized mapping of Appendix~\ref{app:task-contracts} and the arrow its preferred direction ($\uparrow$ higher better, $\downarrow$ lower better). Families divide into three types: \emph{weighted} families combine the listed terms linearly; \emph{gated} families score a revisit-gap or service core and add a post-gate bonus, so the terms enter nonlinearly and no single weight applies; \emph{single-objective} families have a score that is a monotone rescaling of one metric. Cells are penalized five-case means: invalid or missing cases take the score-zero native value, namely zero for higher-is-better metrics, one for the bounded SatNet ratios, the 48-hour horizon for the revisit gap, and the 12-hour horizon for AEOSSP turnaround on cases with no completed task. \emph{valid (/5)} counts verifier-valid cases. $\dagger$ marks count and battery-margin terms reported as valid-only means, since a penalty for a count is ill-defined; $\ddagger$ marks unscored context metrics. Profit is in thousands of objective points, power in kWh, turnaround in seconds, the revisit gap in hours, battery in Wh. Reference columns hold fixed named solvers rather than Table~\ref{tab:main-results}'s per-case envelope: Ref1 and Ref2 are MWIS conflict-graph and greedy-LNS (AEOSSP), $\Delta$-MILP and PPO (SatNet), CP local-search and CELF (Regional), J2-RGT set-cover and RGT-APC constructive (Revisit), MCLP-TEG and UMCF-SRR (Relay), CP local-search and pruned MILP (Stereo), and reference lookup (SPOT-5, one reference). The Relay full-service latency and satellite bonus is gated per case at full service: Codex and Kimi each reach full service on one of the five cases, so their Relay means include that bonus, while the reference rows never reach full service.}
\label{tab:native-metrics}
\end{table*}

Table~\ref{tab:native-metrics} reports the raw mission metrics behind the normalized results in Table~\ref{tab:main-results}, so the mapping in Appendix~\ref{app:task-contracts} can be read against the quantities each source formulation optimizes. The families divide into three scoring types. Three combine several native terms under fixed weights: AEOSSP over completed-weight ratio, completion ratio, turnaround, and power; SatNet over two unsatisfied-demand statistics; Regional over two coverage terms with an action-count and a battery margin. Two are gated rather than linear: Revisit scores a revisit-gap curve and adds a satellite-count bonus only once every target meets its requirement, and Relay scores demand service and worst-group service and adds a latency and satellite bonus only at full service. The last two are single-objective, where the score is a monotone rescaling of one metric: selected profit for SPOT-5 and product quality for Stereo Imaging. A normalized scale is needed where a family folds several mission terms into one comparable number, which is five of the seven families.

SatNet is the sharpest test of whether normalization flatters the agent rows. Under its published unsatisfied-demand metrics, the strongest agent system reaches $U_{\mathrm{rms}}=0.234$ and $U_{\max}=0.519$, improving on both the mixed-integer baseline ($0.302$, $0.673$) and the reinforcement-learning baseline ($0.316$, $0.772$). The normalized scorer records a narrower lead, $69.45$ against the $61.59$ solver-reference row, because it compresses the unsatisfied-demand range onto a bounded scale.

The native columns also locate where the agent rows genuinely trail, on the raw quantity rather than as an effect of the mapping. Regional Coverage separates on its weighted coverage term, where the strongest agent reaches $0.944$ against references at $0.995$ and $0.968$, and where one agent returns valid but empty schedules at zero coverage. That system's nonzero normalized score comes entirely from the action-count and battery-margin terms, so its best-marked zero action count reflects not acting at all rather than better task performance. Stereo Imaging separates on product quality, $0.689$ for the strongest agent against $0.961$ and $0.920$ for the references; because Stereo is single-objective, its native and normalized values agree by construction, so this gap is a real quality deficit and not a normalization artifact.

Read across families, the native columns reproduce the strong-to-weak ordering of Table~\ref{tab:main-results}. The divergences fall on the weighted and gated families, whose score folds in secondary mission terms beyond the leading native metric, such as SatNet's worst-group ratio, the Regional action and battery margins, or the Relay worst-group service. The normalized scale aggregates mission value across these terms rather than relabeling a single metric, which is why it can be read against the native columns without reducing to them.

\subsection{Family-Native Metrics --- Ablations}

The procedure-injection and memory-accumulation ablations are reported here in the same native units as Table~\ref{tab:native-metrics}, for the two families common to both interventions, Regional Coverage and Relay Constellation, and the two OpenCode systems they target. Tables~\ref{tab:skill-injection-appendix} and~\ref{tab:memory-appendix} give each family's primary native metric, weighted coverage ratio for Regional Coverage and demand service fraction for Relay Constellation, as a penalized five-case mean with invalid cases scored zero; both are higher better. The no-procedure and no-memory columns are the same baseline runs as the corresponding rows of Table~\ref{tab:native-metrics}.

\paragraph{Procedure injection.}
The compact procedure gives a short domain workflow and the procedure pack adds broader search and domain guidance. OpenCode + DeepSeek V4 Pro shows the clearest effect on Regional Coverage, where the full pack raises weighted coverage from $0.40$ to $0.64$ while using fewer actions on average than the no-procedure baseline, consistent with a construction loop that broadens candidate strips, ranks them by marginal coverage, and preserves the best verified incumbent. Relay Constellation is more conjunctive, since placement, ground and inter-satellite links, endpoint capacity, simultaneous visibility, and demand-window routing must align before service appears: service rises from $0.56$ to $0.66$ under the compact procedure but does not improve further under the full pack. OpenCode + MiniMax M2.7 gains on both families from a low base, reaching $0.23$ weighted coverage and $0.17$ service, and its worst-served demand window stays at zero throughout, so the procedure widens partial service without closing the hardest demand.

\begin{table*}[!t]
\centering
\footnotesize
\renewcommand{\arraystretch}{1.15}
\begin{tabular*}{\textwidth}{@{\extracolsep{\fill}}lrrr}
\hline
\rowcolor{headerrow}
System & No procedure & Compact & Procedure pack \\
\hline
\rowcolor{headerrow}
\multicolumn{4}{l}{\textbf{Regional Coverage} (weighted coverage ratio $\uparrow$)} \\
\quad OpenCode + DeepSeek V4 Pro & 0.397 & 0.415 & 0.644 \\
\quad OpenCode + MiniMax M2.7 & 0.044 & 0.127 & 0.225 \\
\rowcolor{headerrow}
\multicolumn{4}{l}{\textbf{Relay Constellation} (service fraction $\uparrow$)} \\
\quad OpenCode + DeepSeek V4 Pro & 0.559 & 0.658 & 0.601 \\
\quad OpenCode + MiniMax M2.7 & 0.000 & 0.124 & 0.174 \\
\hline
\end{tabular*}
\caption{Procedure-injection ablation in native units. Entries are penalized five-case means of each family's primary native metric, both higher better, for the compact-procedure and full procedure-pack conditions against the no-procedure baseline. Procedures help most when they turn construction into verifier-aligned candidate generation and incumbent preservation, as in Regional Coverage; in Relay Constellation broader guidance reduces empty-service collapses but does not remove the coupled placement, link-timing, and routing burden.}
\label{tab:skill-injection-appendix}
\end{table*}

\paragraph{Memory accumulation.}
Each memory condition mounts a frozen note set. A donor system (Codex CLI + GPT-5.4, or OpenCode + DeepSeek V4 Pro) accumulated notes over twelve training cases, two per family for the six families with a training split, and the resulting snapshot is mounted unchanged into every evaluated run: no note derives from a held-out test case, and nothing accumulates across test runs. The mounted set always contains notes from all six donor families, so the reported conditions mix same-family training notes with cross-family notes. Memory is more specific than a static procedure and higher variance, since prior-run notes may carry concrete candidate families, warnings from earlier verifier failures, or habits for preserving a valid solution. On Relay Constellation, OpenCode + DeepSeek V4 Pro improves with memory from $0.56$ service with no memory to $0.68$ with Codex-derived notes and $0.83$ with its own prior-run notes, the largest native gain in either ablation. Regional Coverage shows the transfer-mismatch boundary: Codex-derived notes raise weighted coverage to $0.72$, above the $0.65$ from the reader's own notes, so a donor candidate family can transfer better than self-derived memory when it matches the new geometry. OpenCode + MiniMax M2.7 again gains only marginally and stays below $0.18$ on both families; on Relay its DeepSeek-derived condition also drops two of five cases to invalid, so the penalized service of $0.11$ reflects reliability loss rather than quality alone.

\begin{table*}[!t]
\centering
\footnotesize
\renewcommand{\arraystretch}{1.15}
\begin{tabular*}{\textwidth}{@{\extracolsep{\fill}}lrrr}
\hline
\rowcolor{headerrow}
System & No memory & Codex-derived & DeepSeek-derived \\
\hline
\rowcolor{headerrow}
\multicolumn{4}{l}{\textbf{Regional Coverage} (weighted coverage ratio $\uparrow$)} \\
\quad OpenCode + DeepSeek V4 Pro & 0.397 & 0.725 & 0.646 \\
\quad OpenCode + MiniMax M2.7 & 0.044 & 0.097 & 0.175 \\
\rowcolor{headerrow}
\multicolumn{4}{l}{\textbf{Relay Constellation} (service fraction $\uparrow$)} \\
\quad OpenCode + DeepSeek V4 Pro & 0.559 & 0.681 & 0.833 \\
\quad OpenCode + MiniMax M2.7 & 0.000 & 0.043 & 0.110\textsuperscript{\dag} \\
\hline
\end{tabular*}
\caption{Memory-accumulation ablation in native units. Entries are penalized five-case means of each family's primary native metric, comparing no-memory runs with memory accumulated by a Codex donor and by the OpenCode + DeepSeek V4 Pro donor. \textsuperscript{\dag}Two of five cases are invalid in this cell, so the penalized mean substitutes zero for them; the valid-only mean is $0.184$. Memory transfers concrete solver habits but is high variance: a prior candidate family can match the new case and improve search or mismatch it and regress.}
\label{tab:memory-appendix}
\end{table*}

\section{Temporal-Robustness Check}
\label{app:temporal-robustness}

The AEOSSP temporal check asks whether the held-out result is tied to one particular orbital epoch. It compares the default held-out split with a shifted-horizon split built from a later source epoch while keeping the task family and scoring convention fixed. Codex CLI + GPT-5.4 remains close across the two splits, with mean weighted completion ratio changing from 0.7075 to 0.6949. OpenCode + DeepSeek V4 Pro changes from 0.6820 to 0.6816. The two solver references are similarly stable: conflict-graph scheduling changes from 0.7582 to 0.7667, and greedy large-neighborhood search from 0.6816 to 0.6874.

This check isolates one source of temporal sensitivity: the orbital epoch used to generate AEOSSP cases. Under the shifted horizon, both agent systems and solver references preserve their relative behavior, indicating that the AEOSSP comparison is not tied to one frozen time window.

\section{Additional Case Studies}
\label{app:additional-case-studies}

\begin{figure*}[!t]
\centering
\includegraphics[width=\textwidth]{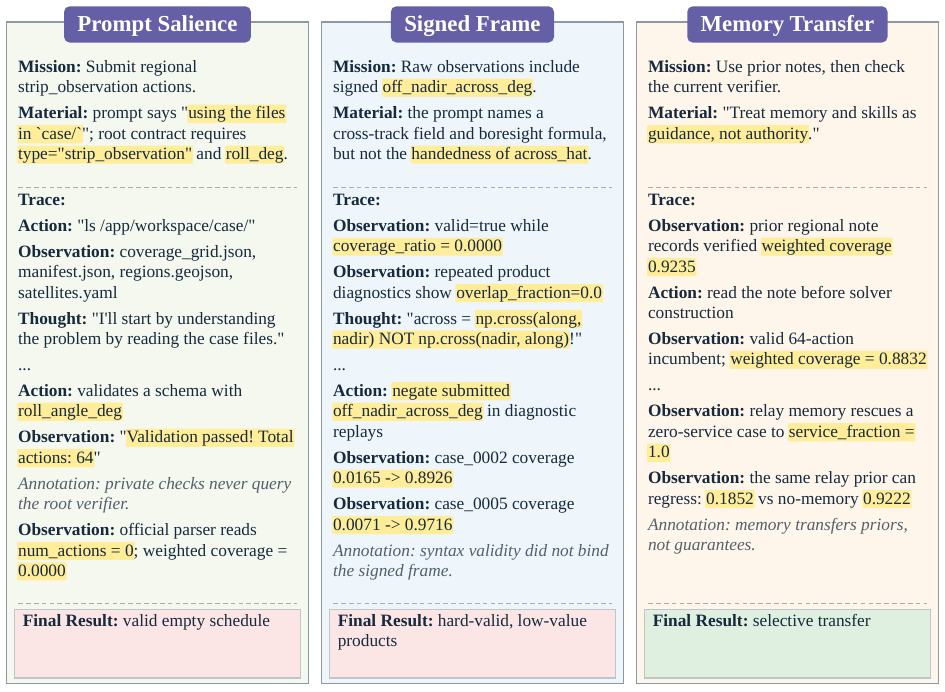}
\caption{Selected Appendix~\ref{app:additional-case-studies} mechanisms connecting agent-facing material, trace behavior, and final outcome. Left: prompt salience leads a Regional Coverage run to bind to case files and an agent-inferred \texttt{roll\_angle\_deg} schema, producing a valid empty schedule under the official parser. Middle: a Stereo Imaging run produces hard-valid actions but loses product overlap because the signed cross-track convention is not bound. Right: memory transfers concrete search priors that help under structural match and regress under mismatch.}
\label{fig:appendix-case-study-materials}
\end{figure*}

This appendix gives trace-level evidence for the mechanisms summarized in Section~\ref{sec:analysis} and Figure~\ref{fig:case-study-traces}. Each example links an agent-facing material choice, such as task wording, workspace packaging, injected procedure, memory note, or final-artifact convention, to an observed run behavior and official outcome. The examples cover prompt salience, signed-frame binding, procedure granularity, memory-prior transfer, and canonical deliverable handoff. Figure~\ref{fig:appendix-case-study-materials} previews the prompt-salience, signed-frame, and memory-transfer examples.

Each case study starts from the official outcome and reconstructs which source the evaluated system treated as defining the task: the short task prompt, root task document, local verifier, injected procedure, memory note, or self-written checker. Prompt fragments are shown before interpretation because the prompt contract is part of the evidence. The interpretations then explain how the observed trace behavior follows from those fragments.

\paragraph{Prompt Salience and Contract Closure.} The Regional Coverage failure starts from schema selection, but the mechanism is more specific than incomplete reading of the task materials. The short tasking cue points the system toward case files, while the task document defines the required action type and field names.
\begin{promptblock}
Please solve the prepared regional strip-coverage case using the files in `case/`.
...
Your job is to produce `solution.json`, a schedule of `strip_observation`
actions that maximizes unique weighted regional coverage while remaining valid.
...
Do not submit your own strip polygons, coverage claims, or access-window
identifiers. Strip geometry and coverage are derived during validation.
...
Actions with another `type` are not strip observations and do not create
coverage; use only `strip_observation` entries.
\end{promptblock}
The trace follows the case-file cue almost literally: it starts with ``using the files in case/'', lists only the case directory, and later checks an agent-invented field named \texttt{roll\_angle\_deg}. Those checks can be internally coherent, and the final JSON can contain many rows, but the official parser recognizes \texttt{strip\_observation} rows with \texttt{roll\_deg}. The aggregate symptom is therefore valid but empty: all five files are valid JSON schedules after parsing, yet the parsed action count and weighted coverage are zero. The mechanism is premature contract closure. After the first durable schema is written, solvers, validators, and progress messages all reinforce that schema; the later search is real work, but it is work inside the wrong contract.

\paragraph{Signed-Frame Binding.} Kimi CLI + Kimi K2.6 shows a different failure in agent-facing material: the action schema is often correct, but the signed local frame is not bound tightly enough before optimization.
\begin{promptblock}
The satellite set is fixed by TLEs in `case/satellites.yaml`; you choose only
raw observation windows and two boresight steering angles.
`off_nadir_along_deg` tilts along the flight direction and
`off_nadir_across_deg` tilts cross-track in the satellite local frame.
The boresight vector is proportional to:
nadir_hat
+ tan(off_nadir_along_deg) * along_hat
+ tan(off_nadir_across_deg) * across_hat
\end{promptblock}
The system reads the task brief and uses the verifier, and the final files can pass action-level checks. The loss comes after parsing: the submitted cross-track steering signs are expressed in the opposite handedness from the verifier's local frame, so image strips land on the wrong side of the ground track. Product overlap then collapses even when convergence and timing look plausible. Diagnostic replays show that negating the across-track sign converts two near-zero schedules into high-coverage schedules. The prompt content is correct but under-specified. A short signed-frame description can still leave the handedness ambiguous for agents writing their own geometry code.

\paragraph{Procedure Granularity and Coupling Burden.} Procedure injection helps unevenly, because a procedure improves a run only when it matches the action space. Unlike the main text's case-adaptive-search mechanism, the mechanism here is the width of the prompt-provided procedure and how much coupled state it asks the agent to coordinate.
\begin{promptblock}
Valid nonzero but weak | Build a small candidate table and add one
fresh-coverage strip at a time. |
...
Build a small candidate table before dense sweeps:
- satellite id
- start time on `manifest.time_step_s`
- duration
- signed roll
...
\end{promptblock}
\begin{promptblock}
If `valid=true` and `service_fraction=0`, the next move is not latency
optimization. The next move is to complete one served path.
Private propagation or visibility code is a candidate generator, not proof.
\end{promptblock}
In Regional Coverage, the full procedure changes the run from recovering basic geometry into broad candidate generation, marginal unique-coverage scoring, local swaps, and verifier-backed best-so-far retention. The result is the largest procedure-injection gain among the tested families for OpenCode + DeepSeek V4 Pro, raising its mean normalized score from 35.79 to 55.25. Relay Constellation is different. It also benefits from service-first reminders, especially avoiding valid zero-service handoffs, but the broader skill pack expands a coupled search over added relays, ground links, inter-satellite links, endpoint caps, satellite link caps, timing, and routed demand windows. The compact procedure has the better mean score because it keeps the run closer to one served path and one verified candidate at a time. The same external guidance helps when it creates a local marginal loop and can hurt when it widens a conjunctive design problem faster than the agent can coordinate it.

\paragraph{Memory Prior Match and Mismatch.} Memory accumulation shows how natural-language context can influence behavior beyond the immediate task brief: the agent reads prior run notes and imports concrete solver habits into the next case.
The useful side is concrete: regional memory can transfer exact candidate-generation recipes and verified thresholds; relay memory can transfer a tested constellation family and a per-sample routing abstraction. This produces large rescues, including a no-memory Relay case with zero service becoming full service under memory. The boundary is equally concrete. The same relay prior can under-serve a different demand-window pattern, and one memory-conditioned case scores far below its no-memory counterpart. Memory is therefore best interpreted as an empirical prior over solver construction. It changes where the search starts and what repair moves are salient; it does not replace case-specific verifier evidence.

\paragraph{Canonical Deliverable Handoff.} Some failures occur after the solver has already found a better artifact. The prompt and shared workspace rules make the canonical final path part of the contract, not merely a convenience. Incumbent preservation concerns retaining a good candidate during search; this mechanism concerns whether that candidate is copied into the collected deliverable before the run ends.
\begin{promptblock}
- Treat `solution.json` as the required final deliverable unless the workspace
  says otherwise.
- The run may be stopped after 2 hours. As soon as you have any valid or
  likely-valid answer, write it to `solution.json` and keep that file valid
  while you continue improving it.
...
- Prefer incremental improvement: preserve the best working `solution.json`
  you have, and only replace it after the replacement is written completely
  and is at least as likely to verify.
\end{promptblock}
Kimi CLI + Kimi K2.6 on Revisit Constellation shows why this matters. The trace can find verified lower-satellite variants while the collected file remains weaker. Another run reaches the revisit floor, starts a riskier lower-satellite search, and terminates with a collected file that loses the primary gate. This is final-state management under a timeout, not weak physical reasoning: each improvement must be copied back before the next risky branch.

\section{Hallucination and Ungrounded-Submission Metrics}
\label{app:hallucination-metrics}

The failure modes in Section~\ref{sec:what-fails} and the case studies in Appendix~\ref{app:additional-case-studies} include hallucination-type behaviors: agents build self-imposed task contracts, optimize hallucinated objectives, trust agent-written validators over the official verifier, or hand off confident but empty artifacts. Table~\ref{tab:hallucination-metrics} quantifies three process-level counterparts over the full main-experiment matrix (175 runs, 35 per system). \emph{Invalid submissions} are runs whose final collected artifact is rejected by the official verifier, typically through fabricated schema fields or malformed artifacts; 22 runs (12.6\%) end this way. \emph{Valid-but-zero submissions} pass the official verifier but score zero: the artifact is real but operationally empty; 16 runs (9.1\%). \emph{Ungrounded commitments} are runs that write a durable candidate or deliverable artifact before reading the root workspace contract, the behavioral precondition for self-imposed task models; 39 runs (22.3\%), and 8 further runs make no detectable durable commitment. These counts key on explicit read events of the root task document; across the recorded traces, no harness auto-injects that document into the model's context, although Codex CLI auto-injects the generic workspace rules file, which is not the task contract.

\begin{table}[h]
\centering
\footnotesize
\begin{tabular}{lrrr}
\hline
System & Invalid & Valid-but-zero & Ungrounded \\
\hline
Claude & 9 & 1 & 17 \\
Codex & 0 & 0 & 1 \\
Kimi & 0 & 4 & 6 \\
OC+MM & 12 & 6 & 15 \\
OC+DS & 1 & 5 & 0 \\
\hline
All & 22 & 16 & 39 \\
\hline
\end{tabular}
\caption{Hallucination-related process metrics over the 175 main-experiment runs (35 per system). Invalid: the final collected artifact is rejected by the official verifier. Valid-but-zero: the artifact verifies but scores zero. Ungrounded: a durable artifact commitment occurs before the root workspace contract is read. Abbreviations follow Table~\ref{tab:main-results}.}
\label{tab:hallucination-metrics}
\end{table}

Two qualifications accompany these counts. First, valid-but-zero submissions have mixed causes: three of the four Kimi CLI + Kimi K2.6 cases on Stereo Imaging stem from an across-track sign-convention mismatch that leaves the artifact valid but with zero footprint overlap, whereas the Relay Constellation cases are zero-service handoffs. Second, an ungrounded commitment is a risk factor, not a verdict: runs can still recover once they acquire the workspace contract and let the official verifier override the private model. Even so, these metrics separate systems whose outcome scores are otherwise close. Codex CLI + GPT-5.4 and OpenCode + DeepSeek V4 Pro almost never commit before reading the contract (1 and 0 of 35 runs, respectively), while Claude Code + Claude Opus 4.6 and OpenCode + MiniMax M2.7 do so in roughly half of their runs, consistent with the trace evidence in Section~\ref{sec:what-fails}.

\end{document}